%% file: freelunch.tex
\documentclass{article}

\PassOptionsToPackage{numbers, compress}{natbib}

\usepackage[preprint]{freelunch}

\usepackage[utf8]{inputenc}
\usepackage[T1]{fontenc}
\usepackage{hyperref}
\usepackage{url}
\usepackage{booktabs}
\usepackage{amsfonts}
\usepackage{nicefrac}
\usepackage{microtype}
\usepackage{xcolor}         
\usepackage{amsmath}
\usepackage{amssymb}
\usepackage{xfrac}
\usepackage{mathtools}
\usepackage{amsthm}
\usepackage{MnSymbol}
\usepackage{xcolor}
\usepackage{colortbl,hhline}
\usepackage[nameinlink]{cleveref}

\usepackage{bm,bbm}
\usepackage{stmaryrd}
\usepackage{booktabs}
\usepackage{color}
\usepackage{tikz}
\usetikzlibrary{patterns}
\usetikzlibrary{external}
\usetikzlibrary{positioning,fit}
\usetikzlibrary{arrows.meta}
\usepackage{pgfplots, pgf, pgffor}
\pgfplotsset{compat=1.18}
\usepgfplotslibrary{groupplots}
\usepgfplotslibrary{statistics}
\usepgfplotslibrary{colorbrewer}
\usepgfplotslibrary{units}
\usepgfplotslibrary{fillbetween}
\usepackage{siunitx}
\usepackage{wrapfig}
\usepackage{caption}
\usepackage{subcaption}
\usepackage{float}
\usepackage{multirow}
\usepackage{multicol}

\pgfplotsset{
legend image code/.code={
\draw[mark repeat=2,mark phase=2]
plot coordinates {
(0cm,0cm)
(0.15cm,0cm)
(0.3cm,0cm)
};%
}
}

\newcommand{\ifstringequal}[4]{%
  \ifnum\pdfstrcmp{#1}{#2}=0
  #3%
  \else
  #4%
  \fi
}

\definecolor{plotcolor1}{HTML}{FFB344}
\definecolor{plotcolor2}{HTML}{E05D5D}
\definecolor{plotcolor3}{HTML}{F283D3}
\definecolor{plotcolor4}{HTML}{5F9EA0}
\definecolor{plotcolor5}{HTML}{417CE7}
\definecolor{plotcolor6}{HTML}{00008B}
\definecolor{plotcolor7}{HTML}{2E8B57}
\definecolor{plotcolor8}{HTML}{8B4513}
\definecolor{plotcolor9}{HTML}{bcbd22}
\definecolor{plotcolor10}{HTML}{17becf}
\pgfplotscreateplotcyclelist{tab10colors}{
    plotcolor1, plotcolor2, plotcolor3, plotcolor5,
    plotcolor4, plotcolor6, plotcolor7, plotcolor8,
    plotcolor9, plotcolor10
}

\colorlet{Xcolor}{plotcolor5!80!black}

\hypersetup{
    colorlinks=true,
    linkcolor={plotcolor4!80!black},
    citecolor={plotcolor4!80!black},
    urlcolor={plotcolor3!80!black}
}

\usepackage{xurl}

\newcommand{\R}{\mathbb{R}}

\newcommand{\norm}[1]{{\left\Vert #1 \right\Vert }}

\usepackage{glossaries}
\glsdisablehyper
\newacronym{imp}{IMP}{Iterative Magnitude Pruning}
\newacronym{peft}{PEFT}{Parameter-Efficient Fine-Tuning}
\newacronym{per}{PERP}{Parameter-Efficient Retraining after Pruning}
\newacronym{ft}{FT}{Fine-Tuning}
\newacronym{lora}{LoRA}{Low-Rank Adaptation}
\newacronym{llm}{LLM}{Large Language Model}
\newacronym{gpt}{GPT}{Generative Pretrained Transformer}
\newacronym{nlp}{NLP}{Natural Language Processing}
\newacronym{bn}{BN}{Batch-Normalization}
\newacronym{ln}{LN}{Layer-Normalization}
\newacronym{multlora}{ScaleLoRA}{ScaleLoRA}
\newacronym{masklora}{MaskLoRA}{MaskLoRA}
\newacronym{lora-prune}{LoRA-Prune}{LoRA-Prune}
\newacronym{nn}{NN}{Neural Network}
\newacronym{obs}{OBS}{Optimal Brain Surgeon}
\newacronym{cnn}{CNN}{Convolutional Neural Network}
\newacronym{dp}{DP}{Dense Propagation}
\newacronym{sp}{SP}{Sparse Propagation}
\newacronym{mp}{MP}{Mixed Propagation}
\newacronym{lm}{LM}{Language Modeling}
\newacronym{mse}{MSE}{Mean Squared Error}
\newacronym{cs}{CS}{Cosine Similarity}

\newcounter{romansubsection}
\renewcommand{\theromansubsection}{(\roman{romansubsection})}

\newcommand{\romansubsection}[1]{%
  \refstepcounter{romansubsection}%
  \subsection*{\theromansubsection\ #1}%
  \addcontentsline{toc}{subsection}{\theromansubsection\ #1}%
}

\title{A Free Lunch in LLM Compression: Revisiting Retraining after Pruning}
\author{
  Moritz Wagner$^1$\thanks{Correspondence to: Moritz Wagner \ \texttt{wagner@zib.de}} \quad Christophe Roux$^{1\ 2}$ \quad Max Zimmer$^{1\ 2}$ \quad Sebastian Pokutta$^{1\ 2}$\\
  $^1$Department for AI in Society, Science, and Technology, Zuse Institute Berlin, Germany\\
  $^2$Institute of Mathematics, Technische Universität Berlin, Germany
}

\begin{document}
\maketitle
\begin{abstract}
Post-training pruning can substantially reduce LLM inference costs, but it often degrades quality unless the remaining weights are adapted. Since global retraining is expensive at LLM scale, recent work has largely focused on increasingly sophisticated pruning criteria that aim to select better sparsity patterns without adaptation. We revisit this trade-off through \emph{local reconstruction}: after pruning, we adapt one subset of the model parameters at a time on a calibration set, training it to match the corresponding intermediate activations of the dense model.
We evaluate local reconstruction across model families and scales, up to 72B parameters, and establish three main findings. First, local reconstruction is an effective adaptation mechanism for LLMs: it matches post-pruning retraining while using over an order of magnitude less data and compute, even when using PEFT techniques. Second, reconstruction exhibits a broad ``free-lunch'' regime in \emph{granularity}, i.e., the reconstruction parameter window: as long as the reconstructed region contains at least a nonlinear submodule, final quality is largely insensitive to the window size, allowing granularity to be chosen primarily based on memory constraints. In contrast, reconstructing individual matrices, despite being the natural approach often proposed in the literature, consistently underperforms, as small matrix-level errors accumulate into larger activation drift. Lastly, reconstruction reduces the relative importance of the pruning criterion: performance gaps between sophisticated criteria and simple baselines shrink with model scale, making simple methods competitive at scale. Overall, our results challenge the prevailing view that post-pruning adaptation is impractical for LLMs.
\end{abstract}

\section{Introduction}

\glspl{llm} have revolutionized natural language processing with state-of-the-art performance across a wide range of tasks from text generation to code synthesis. However, their scale comes with significant computational and memory demands, posing challenges for both researchers and practitioners. Model compression, particularly post-training pruning, addresses these bottlenecks by identifying and removing redundant weights in pretrained \glspl{nn}, yielding \emph{sparse} models with reduced inference costs \citep{Han2015, Gale2019, Hoefler2021}.

Historically, post-training pruning has used a prune-retrain pipeline: remove weights (often by magnitude) and then adapt the remaining weights by retraining to recover accuracy \citep{Han2015, Gale2019, Hoefler2021}. For \glspl{llm}, full retraining is viewed as computationally prohibitive, requiring full-model backpropagation on large data volumes \citep{Sun2023, Frantar2023a, Zimmer2023a}. Without such adaptation, magnitude pruning can cause large quality drops \citep{Sun2023,yin2024owl}, so prior work has largely focused on designing better pruning criteria (rules for deciding which weights to remove) so that the pruned model remains accurate even without post-pruning adaptation.

\begin{figure*}[t]
  \centering
  \input{granularity_figure2.tex}
  \caption{Illustration of reconstruction granularities. At the finest level, \emph{per-matrix} reconstruction adapts each individual weight matrix separately. Block size $\sfrac{1}{2}$ reconstructs the attention and MLP components of each transformer block independently. Block size $1$ reconstructs each full transformer block at a time, and block size $2$ reconstructs two consecutive transformer blocks together.}
  \label{fig:granularity_explained}
\end{figure*}

Such approaches operate \emph{locally}, pruning each weight matrix independently using a small calibration set. Given a single layer with weight matrix $W\in\R^{d_{\text{out}}\times d_{\text{in}}}$ and calibration input $X\in\R^{d_{\text{in}}\times B}$ (obtained by running a small set of sequences containing a total of $B$ tokens through the dense model up to $W$), the per-matrix pruning objective is
\begin{align}
  \min_{M,\hat{W}}\left\| WX-(M\odot \hat{W})X\right\|_F^2,
  \label{eq:layerwise_pruning_bg}
\end{align}
where $\hat{W}\in\R^{d_{\text{out}}\times d_{\text{in}}}$ is the adapted weight matrix, $\odot$ denotes the elementwise Hadamard product, and $M \in \{0,1\}^{d_{\text{out}}\times d_{\text{in}}}$ is a binary pruning mask satisfying some sparsity constraint. Common sparsity constraints are $p$\% unstructured or $N$:$M$ semi-structured sparsity \citep{Mishra2021}, where $N$ out of $M$ contiguous weights are kept. The problem given in \autoref{eq:layerwise_pruning_bg} can be relaxed into two subproblems:
(i) \emph{mask selection} chooses $M$ while keeping the weights $\hat{W} = W$ fixed,
and (ii) \emph{reconstruction} adapts $\hat W$ while keeping $M$ fixed. For instance, Wanda \citep{Sun2023} scores weights by scaling their magnitudes with activation statistics, and does not perform any weight updates, while SparseGPT \citep{Frantar2023a} additionally performs local weight updates to reduce the resulting reconstruction error. A shared premise across many of these approaches is that strong pruning mask selection is the primary lever, because retraining is infeasible. In this work, we focus on the latter problem of reconstruction, noting that reconstructing a single matrix in isolation is a convex quadratic problem, which admits an analytical solution, and requires significantly less memory compared to retraining the entire model.

More precisely, we revisit post-pruning adaptation and study \emph{local reconstruction}, which was first proposed for post-pruning adaptation of \glspl{llm} by \citet{guo2024ebft}. Instead of retraining the full model, we adapt only a small pruned \emph{submodel} (e.g., one transformer block) at a time using a small calibration set by matching intermediate activations of the original dense model.
In this work, we distinguish between \emph{reconstruction} (local adaptation using intermediate activations as targets) and \emph{retraining} (adaptation of the entire pruned model using true labels).

Let $f$ be the submodel we want to reconstruct. The local reconstruction objective is given by
\begin{align}
  \label{eq:functional_reconstruction_bg}
  \min_{\hat\theta}\; \bigl\| f(X;\theta) - f(X; M_\theta \odot \hat\theta) \bigr\|_2^2,
\end{align}
where $\theta$ denotes the dense parameters of the submodel, $\hat\theta$ the adapted parameters, and $M_\theta$ the corresponding pruning mask.
Because $f$ includes nonlinearities and multiple matrices, it can absorb pruning-induced errors locally (e.g., by compensating across projections within attention and MLP blocks), which is not possible when optimizing a single matrix in isolation.

We study a wide range of such submodel sizes, which we refer to as the \emph{granularity} of reconstruction. Our analysis spans from reconstructing individual weight matrices to within-block components (reconstructing attention and MLP separately, which we refer to as ``block size $\sfrac{1}{2}$''), to full transformer blocks (``block size $1$''), and finally to coarser multi-block chunks (cf. \autoref{fig:granularity_explained}). By dissecting local reconstruction at different granularities, we make several surprising findings, which we highlight in detail below: (i) local reconstruction makes post-pruning adaptation feasible in regimes where full retraining or \gls{peft} is not, (ii) we observe a free-lunch regime where reconstructing at smaller granularity reduces peak memory without sacrificing final quality, and (iii) once reconstruction is applied, the apparent advantage of sophisticated pruning criteria shrinks with model size.

\begin{figure*}[t]
  \centering
  \input{ppl_vs_mem_bar_by_model.tex}
  \caption{Perplexity and memory usage of three models pruned to 2:4 sparsity with Wanda and reconstructed with different granularities. The number of calibration samples is 1024. $\sfrac{1}{4}$ of the model is 8, 10, and 16 transformer blocks for LLaMA-3-8B, LLaMA-2-13B, and Qwen-2.5-32B, respectively. We report peak memory as the maximum allocated GPU memory during reconstruction. The compute time is for NVIDIA H100 GPUs. We use as many GPUs as needed to accommodate the reconstruction (at most four GPUs) and do not offload inactive model components.}
  \label{fig:block_size_vs_ppl_mem}
\end{figure*}

\paragraph{Our contributions are summarized as follows.}
We conduct a large-scale computational study to dissect the design choices for local reconstruction, i.e., the post-pruning adaptation of remaining (non-pruned) weights using a small calibration dataset. Our central findings are:

\begin{enumerate}
  \item \textbf{Local reconstruction makes adaptation feasible.} Local reconstruction makes post-pruning adaptation practical at \gls{llm} scale by restricting optimization to a small submodel at a time, so peak memory scales with the reconstructed submodel rather than the full network. We find that local reconstruction matches \gls{lora}-based \gls{peft} retraining while using over an order of magnitude less calibration data and compute, making effective adaptation feasible in regimes where full retraining or \gls{peft} is not.

  \item \textbf{Local reconstruction admits a free-lunch regime.} Intuitively, increasing the reconstruction granularity should trade quality for resources: coarser granularity should improve final model quality but demand more resources for backpropagation, with full retraining being the most resource-intensive and per-matrix reconstruction the least. However, we find that, across a wide practical range, increasing the size of the submodel reconstructed at once does not improve the final model quality. Peak memory usage grows sharply as granularity becomes coarser, while the compute time changes more gradually (cf. \autoref{fig:block_size_vs_ppl_mem}). This yields a free-lunch regime where granularity can be chosen primarily to satisfy hardware constraints, trading memory for compute time without sacrificing final quality. A surprising exception is per-matrix reconstruction, which consistently underperforms despite being the natural and widely used formulation.

  \item \textbf{Mask selection becomes negligible.} With an effective reconstruction step in place, performance gaps between sophisticated pruning criteria and simple baselines shrink with model size, making simple methods increasingly competitive. This is surprising given prior evidence that magnitude pruning is prone to failure in \glspl{llm}. Reconstruction largely mitigates this failure, suggesting that part of the apparent advantage of sophisticated mask selection comes from compensating for missing or weak post-pruning adaptation.
\end{enumerate}

\noindent Taken together, these findings challenge the prevailing assumption that local post-pruning adaptation is insufficient to recover pruning-induced performance degradation. This renders the prune-retrain pipeline, restricted to a local variant, a practical alternative to skipping adaptation entirely at \gls{llm} scale. Our results further suggest that much of the complexity in recent pruning methods may partially compensate for missing adaptation rather than address fundamental limitations of simpler approaches.

\section{Background}
\label{sec:related_work}

\begin{table*}[t]
  \caption{Comparison of local reconstruction (block size 1) and MaskLoRA fine-tuning. The number of calibration samples for pruning is 1024 and the pruning method is Wanda in all cases. $\downarrow$: lower is better, $\uparrow$: higher is better.\newline}
  \scriptsize
  \setlength{\tabcolsep}{3pt}
      \centering
      \begin{tabular}{cccccccc}
      \toprule
      Model & Sparsity Type & Method & Number of Samples & Perplexity $\downarrow$ & Accuracy $\uparrow$ & Compute Time & GPU \\
      \midrule
      \multirow{3}{*}{LLaMA-3-8B} & \multirow{3}{*}{50\%} & Prune only & - & 9.14 & 56.61\% & 3 minutes & H200 \\
      & & Local reconstruction & 2048 & 7.68 & 58.22\% & 1.4 hours & H200 \\
      & & MaskLoRA fine-tuning & 131,072 & 7.70 & 58.13\% & 19 hours & H200 \\
      \midrule
      \multirow{3}{*}{\begin{tabular}{c}Qwen-2.5-7B\\-Instruct\end{tabular}} & \multirow{3}{*}{2:4} & Prune only & - & 16.01 & 59.64\% & 3 minutes & A100 80GB \\
      & & Local reconstruction & 2048 & 8.93 & 64.16\% & 1.5 hours & A100 80GB \\
      & & MaskLoRA fine-tuning & 131,072 & 8.89 & 64.02\% & 196 hours & A100 80GB \\
      \bottomrule
      \end{tabular}
      \label{tab:mask_lora_vs_local_rec}
\end{table*}

Post-training pruning is often studied in a prune-retrain setting: weights are removed and the remaining ones are adapted to recover accuracy. Early work emphasizes iterative procedures and second-order criteria such as Optimal Brain Surgeon (OBS) \citep{LeCun1989,Hassibi1992}, and later work emphasizes iterative prune-retrain schemes for reaching high sparsity and uncovering winning ticket subnetworks \citep{Han2015,Frankle2018}. For \glspl{llm}, full retraining after pruning is considered infeasible in practice due to their large memory and compute footprints and the need for large datasets for retraining. Consequently, much recent work targets settings where post-pruning adaptation is either unnecessary or minimal, making the pruning criterion itself the primary lever. In this regime, magnitude pruning is prone to failure for \glspl{llm}. This failure is linked to distinctive activation and weight statistics in \glspl{llm} compared to smaller deep learning models such as convolutional \glspl{nn} \citep{xiao2023smoothquant,Sun2023,yin2024owl}. \citet{dettmers2022llmint88bitmatrixmultiplication} find that \glspl{llm} exhibit outlier features, i.e., features that have significantly larger magnitudes than the other features in the same layer. Intuitively, outlier activations cause a few input directions to dominate a layer's output, so pruning weights purely by magnitude can remove parameters that are small in absolute value but crucial for preserving those high-activation features, leading to large errors after pruning. This motivates the development of more sophisticated pruning methods that do not rely on full retraining.

Many \gls{llm} pruning methods therefore operate \emph{locally} at the level of individual weight matrices using a small calibration set \citep{Sun2023,zhang2024plug,Frantar2023a}. One line of work focuses on mask selection while keeping the remaining weights fixed: Wanda scores weights using activation statistics \citep{Sun2023}, and RIA further accounts for relative neuron importance \citep{zhang2024plug}. A second line \emph{interleaves} mask selection with \emph{per-matrix reconstruction}, updating the surviving weights to reduce activation mismatch on the calibration set. SparseGPT \citep{Frantar2023a} is a well-established approach in this category, combining OBS-style selection with linear least squares updates. Related methods explore alternative optimization strategies for similar per-layer objectives \citep{zhao2024convex,bovzafast}.

Complementary work emphasizes \emph{post-pruning adaptation} by explicitly updating the remaining weights \emph{after} pruning. The most local variant performs per-matrix reconstruction (with the mask fixed) by minimizing activation error for each linear layer independently as in \autoref{eq:layerwise_pruning_bg} \citep{wang2024reconstruct}. More recent methods reconstruct larger transformer submodels by matching intermediate activations of the dense model as in \autoref{eq:functional_reconstruction_bg}, enabling local error compensation across coupled projections and nonlinearities \citep{guo2024ebft,Shin2024}. Related results for quantized vision models suggest that optimizing smaller submodels can outperform full-model calibration under limited calibration data \citep{li2021brecq,jeon2022mr}. 

Another alternative is \gls{peft} after pruning, where a small set of trainable parameters is introduced or selected while most pretrained weights remain frozen \citep{Zimmer2023a,Munoz2024}. \gls{lora} is a common \gls{peft} mechanism \citep{Hu2021b}, and MaskLoRA keeps adapters sparse so they can be merged without sacrificing sparsity \citep{Zimmer2023a}. Related work also studies evolving the pruning mask during fine-tuning \citep{xiao2025leave}. See \autoref{app:additional_related_work_notes} for further related work.

Several techniques reduce the memory footprint of fine-tuning without changing the objective, such as activation checkpointing (trading compute for lower activation memory), microbatching with gradient accumulation (reducing peak activations), and \gls{peft} methods such as \gls{lora} (reducing optimizer state and gradient memory but still requiring end-to-end backpropagation through the full model). Local reconstruction is complementary and strictly more modular: by restricting optimization to a small submodel at a time, it allows reconstructing very large models by only loading the active submodel (and its optimizer state) on GPU, while keeping the rest of the network inactive.

\section{Experiments}
\label{sec:experiments}

\begin{figure}
  \centering
  \begin{minipage}[t]{.48\textwidth}\vspace{0pt}
    \centering
    \input{ppl_vs_numsamples_only.tex}
    \captionof{figure}{Perplexity of OPT-1.3B pruned to 50\% unstructured sparsity with Wanda and reconstructed at different calibration set sizes. The shaded areas indicate the min-max range over random seeds. Reconstructing the full decoder is block size 24.}
    \label{fig:ppl_vs_numsamples}
  \end{minipage}\hfill
  \begin{minipage}[t]{.48\textwidth}\vspace{0pt}
    \centering
    \input{lr_figure2.tex}
    \vspace{0pt}
    \captionof{figure}{Average zero-shot accuracy of Qwen-2.5-7B-Instruct pruned to 2:4 sparsity and reconstructed at different learning rates. The number of calibration samples for pruning is 1024 and the pruning method is Wanda. The shaded areas indicate the min-max range over random seeds.}
    \label{fig:qwen_lrs}
  \end{minipage}
\end{figure}

We study post-training pruning and post-pruning adaptation on models from the OPT, LLaMA-2, LLaMA-3, and Qwen-2.5-Instruct families \citep{Zhang2022b,Touvron2023,Grattafiori2024,qwen2}.
For pruning and reconstruction, we use calibration data sampled from C4 \citep{Raffel2020}. Each sample is a tokenized text sequence truncated to a length of $L=\min(L_{\text{model}},4096)$ tokens, where $L_{\text{model}}$ is the model's maximum context length. We consider both unstructured sparsity and $N\!:\!M$ semi-structured patterns.
All linear layers except the embedding and LM head are pruned with uniform sparsity following \citet{Sun2023}.
We evaluate perplexity on WikiText-2 \citep{merity2016pointer} and average zero-shot accuracy using the EleutherAI evaluation harness \citep{eval-harness}. The zero-shot accuracy is averaged over the tasks Winogrande \citep{Sakaguchi2021}, GLUE RTE \citep{Wang2018d}, OpenBookQA \citep{Mihaylov2018}, BoolQ \citep{Clark2019}, HellaSwag \citep{Zellers2019}, ARC-Easy, and ARC-Challenge \citep{Clark2018}. Our code is available at \href{https://github.com/ZIB-IOL/Free-Lunch-in-LLM-Compression}{github.com/ZIB-IOL/Free-Lunch-in-LLM-Compression}.

We optimize the reconstruction loss given in \autoref{eq:functional_reconstruction_bg} using AdamW with a linear learning rate schedule, a warmup period of 10\% of the total optimization steps, and a batch size of 2.
To compare granularities, we fix the number of epochs per reconstructed submodel, so that each parameter is optimized for the same number of minibatch updates across settings. Changing granularity primarily changes which parameters are optimized jointly, rather than how many times any given parameter is updated. For specific hyperparameters, see \autoref{app:hyperparameters}.

There are several design choices regarding the reconstruction objective, including the propagation strategy (how calibration data defines inputs and targets) and the loss function (how reconstruction quality is measured). We use MSE as the loss function with activations from the pruned prefix as inputs and activations from the dense model as targets as in \autoref{eq:functional_reconstruction_bg}. Ablations over alternative reconstruction variants using different loss functions and propagation strategies (both inputs and targets from the pruned or dense model) are deferred to \autoref{app:additional_experiments_loss_and_propagation}. We find that the choice of loss function and propagation strategy has no significant impact on the final model quality. Furthermore, \autoref{app:additional_experiments_minipile} presents additional experiments investigating the effect of the calibration set on the reconstruction quality and extends evaluation to different benchmarks, and \autoref{app:structured_pruning} shows how reconstruction affects structured pruning and higher sparsity levels.

Our experiments are organized around four questions:
(i) How does local reconstruction compare to retraining (with \gls{peft}) in terms of compute time and hardware requirements?
(ii) How does per-matrix reconstruction perform relative to coarser submodel-wise local reconstruction?
(iii) Does submodel reconstruction granularity affect final model quality?
(iv) After effective reconstruction, how important is the pruning criterion?

\romansubsection{How does reconstruction compare to retraining?}
\label{sec:experiments:masklora}

A common approach to post-pruning adaptation is retraining the model end-to-end. Since full retraining is prohibitively expensive for \glspl{llm}, we apply \gls{peft} to pruned models above 1.3 billion parameters, specifically the sparsity-preserving LoRA variant MaskLoRA \citep{Zimmer2023a}, and only fully fine-tune OPT-1.3B.
\autoref{tab:mask_lora_vs_local_rec} compares local reconstruction (block size $1$) to MaskLoRA fine-tuning under two representative settings. 
On both LLaMA-3-8B and Qwen-2.5-7B-Instruct, local reconstruction achieves essentially the same perplexity and accuracy as MaskLoRA while using 64$\times$ fewer samples. For LLaMA-3-8B, this leads to a 13.6$\times$ speedup, while for Qwen-2.5-7B-Instruct, it leads to a 130$\times$ speedup. \autoref{fig:ppl_vs_numsamples} shows that, at equal calibration set sizes, local reconstruction achieves much lower perplexity than full fine-tuning on OPT-1.3B. Even when increasing the calibration set size to 131k samples, full fine-tuning OPT-1.3B yields higher perplexity than reconstruction with block size $\sfrac{1}{2}$ and 1024 samples (15.85 vs. 15.60 perplexity).

These results underscore a practical advantage of local reconstruction: \textbf{Fine-tuning a pruned model end-to-end requires orders of magnitude more data and hence substantially more compute to achieve the same final model performance.} Moreover, local reconstruction can be performed in a memory-constrained setting. For example, a 32B model can be locally reconstructed on a single 80GB H100 without gradient checkpointing, offloading inactive model components, or quantization. Sparsity-preserving LoRA variants such as MaskLoRA, on the other hand, need the masked low-rank adapter matrices for the backward pass and hence require gradient checkpointing when applied to larger models (without gradient checkpointing, MaskLoRA fine-tuning of Qwen-2.5-32B with the settings detailed in \autoref{app:hyperparameters} requires 275GB of GPU memory).

\romansubsection{How does per-matrix reconstruction compare?}
\label{sec:experiments:per_matrix}

\begin{figure}
  \centering
  \begin{minipage}[t]{.48\textwidth}\vspace{0pt}
    \centering
    \input{error_plot.tex}
  \captionof{figure}{Accumulated reconstruction error (MSE) normalized by dense activation norm after attention key-projection layers for different reconstructed models. The base model is LLaMA-3-8B pruned to 2:4 sparsity with Wanda. The calibration set size is 1024.}
  \label{fig:error_plot}
  \end{minipage}\hfill
  \begin{minipage}[t]{.48\textwidth}\vspace{0pt}
    \centering
    \input{error_plot_local.tex}
  \captionof{figure}{Matrix-local reconstruction error (MSE) normalized by dense activation norm after attention key-projection layers for different reconstructed models. The base model is LLaMA-3-8B pruned to 2:4 sparsity with Wanda. The calibration set size is 1024.}
  \label{fig:error_plot_local}
  \end{minipage}
\end{figure}

\begin{figure}
  \centering
  \begin{minipage}[t]{.48\textwidth}\vspace{0pt}
    \centering
    \input{error_hist.tex}
  \captionof{figure}{Histogram of normalized MSE between activations after key-projection layers in differently reconstructed models. The counts are normalized such that both plots integrate to 1. The base is LLaMA-3-8B pruned to 2:4 sparsity with Wanda. The calibration set size is 1024.}
  \label{fig:error_hist}
  \end{minipage}\hfill
  \begin{minipage}[t]{.48\textwidth}\vspace{0pt}
    \centering
    \input{acc_vs_scale.tex}
  \captionof{figure}{Average zero-shot error rate (1 - accuracy) vs. model size for Qwen-2.5 Instruct models when pruned to 2:4 sparsity. The number of calibration samples is 256. Both axes are on a log scale.}
  \label{fig:acc_vs_scale}
  \end{minipage}
\end{figure}

\begin{table}[t]
  \caption{Per-matrix vs. block size 1 local reconstruction for multiple models pruned to 2:4 and 60\% unstructured sparsity. The number of calibration samples is 256 and the pruning method is Wanda. The best result for each model and sparsity type combination is highlighted in bold. $\downarrow$: lower is better, $\uparrow$: higher is better.\newline}
  \label{tab:per_matrix_underperforms}
  \scriptsize
  \centering
  \setlength{\tabcolsep}{5pt}
  \begin{tabular}{l c cccccccc}
  \toprule
  \multirow{4}{*}{Model} & \multicolumn{4}{c}{2:4 Sparsity} & \multicolumn{4}{c}{60\% Sparsity}\\
  \cmidrule(lr){2-5} \cmidrule(lr){6-9}
  & \multicolumn{2}{c}{Accuracy (in \%) $\uparrow$} & \multicolumn{2}{c}{Perplexity $\downarrow$} & \multicolumn{2}{c}{Accuracy (in \%) $\uparrow$} & \multicolumn{2}{c}{Perplexity $\downarrow$} \\
   \cmidrule(lr){2-3} \cmidrule(lr){4-5} \cmidrule(lr){6-7} \cmidrule(lr){8-9}
  &
  Per-matrix & Block size 1 & Per-matrix & Block size 1 &
  Per-matrix & Block size 1 & Per-matrix & Block size 1 \\
  \midrule
  OPT-6.7B     & 47.02 & \textbf{48.98} & 14.25 & \textbf{12.70} & 44.31 & \textbf{46.94} & 15.76 & \textbf{13.22} \\
  LLaMA-2-13B  & 54.15 & \textbf{56.49} & 7.59 & \textbf{6.22} & 56.43 & \textbf{58.02} & 7.22 & \textbf{6.21} \\
  LLaMA-2-70B  & 63.14 & \textbf{63.84} & 5.01 & \textbf{4.56} & 63.81 & \textbf{63.98} & 4.72 & \textbf{4.52} \\
  Qwen-2.5-14B & 64.49 & \textbf{65.68} & 58.58 & \textbf{12.85} & 64.66 & \textbf{65.30} & 76.10 & \textbf{9.23} \\
  Qwen-2.5-32B & 67.10 & \textbf{67.39} & 13.89 & \textbf{6.50} & 66.29 & \textbf{67.30} & 7.02 & \textbf{6.28} \\
  Qwen-2.5-72B & \textbf{69.99} & 69.94 & 5.96 & \textbf{5.86} & 69.78 & \textbf{70.18} & \textbf{5.68} & 5.70 \\
  \bottomrule
  \end{tabular}
\end{table}

\autoref{tab:per_matrix_underperforms} compares per-matrix reconstruction to local reconstruction at the granularity of full transformer blocks (block size $1$) when pruning with Wanda.
Overall, the pattern is consistent: \textbf{Across models and sparsities, per-matrix reconstruction underperforms compared to block-wise reconstruction.} It almost always yields worse perplexity and zero-shot accuracy than block-wise reconstruction, often by a clear margin. This trend holds across additional models and pruning methods beyond Wanda. The corresponding results are reported in \autoref{app:additional_experiments_per_matrix}. Interestingly, the gap shrinks with model scale and becomes negligible at 72B parameters, where per-matrix reconstruction can serve as a viable, more memory-efficient alternative.

To understand why per-matrix reconstruction underperforms, we distinguish the \emph{matrix-local error} (output discrepancy when dense and reconstructed weights receive the same dense-model inputs) from the \emph{accumulated error} (activation mismatch when the reconstructed model runs normally). \autoref{fig:error_plot}, \autoref{fig:error_plot_local}, and \autoref{fig:error_hist} report both errors after every transformer block's attention key-projection layer. The qualitative conclusions are not specific to this choice. All granularities achieve similar matrix-local errors (\autoref{fig:error_plot_local}), so per-matrix reconstruction does not fail to fit individual matrices. However, it induces substantially larger accumulated error (\autoref{fig:error_plot}), while all coarser granularities, already from block size $\sfrac{1}{2}$, produce similar, much smaller errors.

To test whether the benefit of coarser reconstruction stems from jointly optimizing architecturally coherent submodules (e.g., attention or MLP) or more generally from the presence of nonlinearities, we reconstruct each transformer block in three chunks that deliberately mix attention and MLP projections (chunk~1: \texttt{k\_proj}, \texttt{up\_proj}; chunk~2: \texttt{q\_proj}, \texttt{gate\_proj}; chunk~3: \texttt{v\_proj}, \texttt{o\_proj}, \texttt{down\_proj}), each containing a nonlinearity and optimized on the output of its last projection. On LLaMA-3-8B with 2:4 Wanda pruning, this mixed grouping achieves 12.1 perplexity, substantially better than per-matrix reconstruction (19.2) and close to block size $\sfrac{1}{2}$ (10.7), confirming that the critical factor is the inclusion of a nonlinearity in the reconstructed submodel, not the specific architectural grouping.

The coarse methods are similar not only in error magnitude but also in the activations they produce. For each pair of reconstruction granularities (e.g., block size $\sfrac{1}{2}$ and block size $1$), we compute the normalized MSE between the intermediate activations of the two resulting models at every key-projection layer. \autoref{fig:error_hist} shows histograms of these distances. When comparing two coarse-grained (every granularity except per-matrix) models to each other, the distances concentrate near zero, indicating that they arrive at nearly the same internal representations. In contrast, comparing any coarse model to the per-matrix model yields substantially larger distances. This suggests that the coarse methods converge to a common solution class, whereas per-matrix reconstruction reaches a qualitatively different one.

Taken together, the limitation of per-matrix reconstruction is fundamentally \emph{compositional}: matching each linear layer on dense activations does not ensure the composed sequence of matrices and nonlinearities stays close to the dense computation. Larger nonlinear submodels provide the joint flexibility to absorb these perturbations, yielding lower accumulated error and better downstream performance.

\romansubsection{Does granularity affect final model quality?}
\label{sec:experiments:granularity}

\begin{table}[t]
  \caption{Perplexity and zero-shot accuracy of three different models pruned to 50\% unstructured and 2:4 sparsity and reconstructed at different granularities. The number of calibration samples for pruning is 1024 and the pruning method is Wanda. The results are averaged over multiple random seeds. The best result for each model and sparsity type combination is highlighted in bold. $\downarrow$: lower is better, $\uparrow$: higher is better. B$x$: block size $x$, D$\sfrac{1}{4}$: quarter of the decoder reconstructed at once (block size 8, 10, and 16 for LLaMA-3-8B, LLaMA-2-13B, and Qwen-2.5-32B, respectively).\newline}
  \scriptsize
  \setlength{\tabcolsep}{6.5pt}
  \centering
  \def\vspacemodelcell{-5pt}
  \begin{tabular}{cccccccccccc}
  \toprule
  \multirow{2}{*}{\vspace{\vspacemodelcell}\begin{tabular}{c}Model\\Dense PPL\end{tabular}} & \multirow{2}{*}{Sparsity} & \multicolumn{5}{c}{ Accuracy (in \%) $\uparrow$} & \multicolumn{5}{c}{Perplexity $\downarrow$} \\ 
   \cmidrule(lr){3-7} \cmidrule(lr){8-12}
   & & No rec. & B$\sfrac{1}{2}$ & B1 & B2 & D$\sfrac{1}{4}$ & No rec. & B$\sfrac{1}{2}$ & B1 & B2 & D$\sfrac{1}{4}$ \\
  \midrule
  LLaMA-3-8B & 50\% & 56.89 & \textbf{59.77} & 58.67 & 58.36 & 57.90 & 8.96 & 7.97 & \textbf{7.72} & 7.75 & 7.89 \\
  5.83 & 2:4 & 45.67 & \textbf{55.23} & 54.44 & 52.72 & 52.87 & 21.85 & \textbf{10.24} & 10.31 & 10.32 & 10.53 \\
  \midrule
  LLaMA-2-13B & 50\% & 60.25 & 61.14 & \textbf{61.43} & 60.86 & 61.06 & 5.54 & \textbf{5.25} & \textbf{5.25} & 5.27 & 5.28 \\
  4.57 & 2:4 & 53.23 & \textbf{58.77} & 58.65 & 57.84 & 57.11 & 8.39 & \textbf{6.22} & 6.29 & 6.35 & 6.32 \\
  \midrule
  Qwen-2.5-32B & 50\% & 68.03 & 69.05 & \textbf{69.18} & 69.15 & 69.09 & 6.09 & 5.93 & \textbf{5.92} & 5.98 & 5.97 \\
  4.90 & 2:4 & 66.41 & 68.79 & 69.06 & \textbf{69.19} & 69.05 & 8.00 & 6.06 & \textbf{5.97} & 6.01 & 6.13 \\
  \bottomrule
  \end{tabular}
\label{tab:allmodelsMP_reconstruction}
\end{table}

\autoref{tab:allmodelsMP_reconstruction} compares reconstruction at different granularities for multiple models and sparsity types. From separate reconstruction of attention and MLP components to reconstruction of a quarter of the model at once, perplexity and zero-shot accuracy are very similar and the differences are small relative to the overall prune-to-reconstruct gains. For LLaMA-3-8B and OPT-1.3B (\autoref{fig:ppl_vs_numsamples}), more coarse-grained reconstruction even tends to perform worse. This trend coincides with the findings of \citet{li2021brecq} that more coarse-grained local reconstruction of quantized vision models tends to perform worse. However, we find that this effect does not continue to hold for models above 8 billion parameters. More detailed results are reported in \autoref{app:additional_experiments_per_matrix}.

Taken together, these results reveal a surprising insight: one might expect coarser reconstruction, and ultimately full end-to-end fine-tuning, to yield the best quality, since it can coordinate updates across more parameters and nonlinearities (albeit at higher compute and memory cost). However, this is not what we observe. \textbf{The final model quality is largely insensitive to granularity.} An exception is per-matrix reconstruction, which consistently underperforms (cf. (ii)). Hence, granularity should be chosen primarily for feasibility: since quality is largely insensitive to how much of the model is reconstructed at once, we can trade memory for compute by using smaller reconstructed submodels. \autoref{fig:block_size_vs_ppl_mem} shows how peak memory usage increases sharply with granularity while the final perplexity stays flat, whereas compute time changes more gradually with granularity. This creates a free-lunch scenario where we can choose a granularity that enables reconstruction on given hardware while model quality is not impacted by this choice; the only trade-off is compute time. The overhead of storing target activations is small compared to the overall memory savings of local reconstruction and does not depend on the reconstruction granularity. For example, for Qwen-2.5-32B, block size $\sfrac{1}{2}$ reconstruction saves 247GB of VRAM compared to block size 16 while the overhead of storing target activations is only 20GB of CPU memory. In practice, the learning-rate choice is similarly stable across most granularities, and only needs a slightly smaller value for very coarse-grained reconstruction (see \autoref{fig:qwen_lrs}).

\romansubsection{How important is the pruning criterion after reconstruction?}
\label{sec:experiments:mask_vs_reconstruction}

We now analyze how much the pruning criterion matters once local reconstruction is applied.
\autoref{tab:mag_scale_competitive} highlights a premise that has shaped much recent \gls{llm} pruning work: if post-pruning adaptation is assumed infeasible, then mask selection is the primary lever, and we indeed observe large gaps between magnitude pruning and stronger criteria such as Wanda and SparseGPT in the \emph{No Reconstruction} columns. This explains why increasingly sophisticated mask-selection rules have been a central focus: without reconstruction, their effect is large. 
The \emph{Reconstruction} columns show a different story. \textbf{Once we \emph{do} allow a post-pruning adaptation step, much of the pruning-induced error can be absorbed, and the performance differences between pruning criteria disappear with increasing model size.} This reduced dependence on the pruning criterion is especially pronounced at larger scales. \autoref{fig:acc_vs_scale} visualizes that reconstruction largely closes the accuracy gap between magnitude pruning and more sophisticated criteria as model size increases, and in some cases magnitude pruning even matches or slightly surpasses SparseGPT after reconstruction. Additionally, \autoref{fig:acc_vs_scale} shows that Wanda and SparseGPT are essentially equivalent in terms of zero-shot accuracy after reconstruction, while there is a clear gap without reconstruction.

Overall, these results suggest that part of the apparent advantage of sophisticated pruning criteria without reconstruction comes from compensating for missing post-pruning adaptation. Once a strong reconstruction step is in place, the choice of pruning criterion becomes a secondary consideration at scale.

\begin{table*}[t]
  \caption{Magnitude pruning vs. Wanda and SparseGPT. The models are pruned to 2:4 sparsity and reconstructed with block size 1 and 256 calibration samples. The best result for each model is highlighted in bold. $\downarrow$: lower is better, $\uparrow$: higher is better.\newline}
  \label{tab:mag_scale_competitive}
  \scriptsize
  \setlength{\tabcolsep}{4.8pt}
  \centering
  \begin{tabular}{l cccccccccccc}
  \toprule
   & \multicolumn{6}{c}{Accuracy (in \%) $\uparrow$} & \multicolumn{6}{c}{Perplexity $\downarrow$} \\
  \cmidrule(lr){2-7} \cmidrule(lr){8-13}
   & \multicolumn{3}{c}{No Reconstruction} & \multicolumn{3}{c}{Reconstruction} & \multicolumn{3}{c}{No Reconstruction} & \multicolumn{3}{c}{Reconstruction} \\
  \cmidrule(lr){2-4} \cmidrule(lr){5-7} \cmidrule(lr){8-10} \cmidrule(lr){11-13}
  Model & Mag. & Wanda & S.GPT & Mag. & Wanda & S.GPT & Mag. & Wanda & S.GPT & Mag. & Wanda & S.GPT \\
  \midrule
  Qwen-2.5-0.5B & 33.28 & 36.76 & 38.53 & 39.25 & \textbf{43.08} & 42.85 & 21017.80 & 65.86 & 28.16 & 31.35 & 20.48 & \textbf{19.54} \\
  Qwen-2.5-3B   & 32.28 & 48.31 & 52.90 & 51.31 & 55.29 & \textbf{55.38} & 27844.06 & 22.10 & 16.17 & 23.07 & \textbf{12.59} & 12.92 \\
  LLaMA-2-7B & 47.47 & 48.61 & 51.57 & 49.02 & 52.36 & \textbf{52.97} & 36.09 & 9.90 & 9.14 & 10.06 & \textbf{6.98} & 7.01 \\
  Qwen-2.5-7B   & 41.23 & 59.18 & 60.81 & 61.52 & 61.52 & \textbf{61.63} & 5974.84 & 15.98 & 13.02 & 10.18 & \textbf{9.60} & 9.69 \\
  LLaMA-2-13B & 49.85 & 53.67 & 55.92 & 53.46 & 56.49 & \textbf{57.86} & 7.37 & 7.71 & 7.62 & 7.04 & \textbf{6.22} & 6.39\\
  Qwen-2.5-14B  & 53.83 & 63.75 & 64.76 & 65.02 & \textbf{65.68} & 65.63 & 1424.70 & 242.79 & 198.16 & 12.11 & 12.85 & \textbf{10.87} \\
  Qwen-2.5-32B  & 58.55 & 66.41 & 66.33 & 67.32 & \textbf{67.39} & 67.26 & 38.34   & 37.52  & 7.11  & \textbf{6.31} & 6.50 & 6.71 \\
  LLaMA-2-70B & 58.87 & 62.86 & 62.47 & 63.12 & \textbf{63.84} & 63.67 & 6.23 & 5.02 & 5.22 & 4.60 & \textbf{4.56} & 4.79 \\
  Qwen-2.5-72B  & 51.07 & 69.31 & 69.20 & 69.03 & \textbf{69.94} & 69.44 & 226.30  & 5.99  & 6.83  & 5.93 & \textbf{5.86} & 6.39 \\
  \bottomrule
  \end{tabular}
\end{table*}

\section{Conclusion}
\label{sec:conclusion}

Within the scope of our study, which we discuss in detail, together with the
broader impact of our work, in \autoref{app:limitations} and
\autoref{app:impact_statement}, our results support three main findings.

First, local reconstruction is not merely a cheap approximation to end-to-end fine-tuning: it matches and sometimes surpasses full fine-tuning and LoRA-style baselines, while operating in a substantially cheaper data and compute regime.
Second, reconstruction granularity admits a broad free-lunch regime: beyond per-matrix reconstruction (which is consistently worse), final perplexity and zero-shot accuracy are largely insensitive to how much of the model is reconstructed at once, even though peak memory varies sharply. Granularity should therefore be chosen primarily based on hardware constraints rather than expected quality gains.
Third, effective reconstruction changes the role of mask selection: after reconstruction, performance gaps between pruning criteria disappear with increasing model scale, making simple baselines increasingly competitive at larger sizes.

Together, these findings challenge the ``avoid retraining at all costs'' narrative in \gls{llm} pruning. When done properly, local post-pruning reconstruction is not a second-best fallback but a strong default: it can rival PEFT and full fine-tuning while being feasible at model scales where these end-to-end adaptation methods are simply too expensive, and it shifts the focus from more elaborate mask-selection rules toward efficient, well-configured reconstruction.

\section*{Acknowledgements}
This research was partially supported by the DFG Cluster of Excellence MATH+ (EXC-2046/2, project id 390685689) funded by the Deutsche Forschungsgemeinschaft (DFG) as well as by the German Federal Ministry of Research, Technology and Space (research campus Modal, fund number 05M14ZAM, 05M20ZBM) and the VDI/VDE Innovation + Technik GmbH (fund number 16IS23025B).

\newpage

\input{freelunch.bbl}
\newpage
\appendix

\section{Additional Related Work Notes}
\label{app:additional_related_work_notes}

Two adjacent directions are worth noting.
First, several methods study \emph{where} to place sparsity across a network (e.g., non-uniform sparsity allocation across layers/blocks) under a fixed global budget \citep{xubesa,wu2024llm,an2024fluctuation,yin2024owl}.
Second, sparse training methods maintain sparsity throughout training by dynamically updating connectivity patterns \citep{Bellec2017,Mocanu2018,Mostafa2019,Dettmers2019,Evci2019,Liu2020}. These methods study different objectives and settings than the ones of post-training local reconstruction and introduce additional confounders (e.g., varying where sparsity is placed), so we leave them out of scope and focus on reconstruction after post-training pruning.

\section{Limitations}
\label{app:limitations}

Our study is empirical, and its conclusions should be read within the scope of the setting we evaluate. All experiments are conducted on dense decoder-only transformer LLMs with calibration data mainly drawn from C4 (with additional corpora in \autoref{app:additional_experiments_minipile}) and evaluation based on standard perplexity and zero-shot benchmarks (together with a limited set of reasoning probes in \autoref{app:additional_experiments_minipile}). Within this scope, our findings on granularity, on the role of the pruning criterion, and on the competitiveness of local reconstruction are consistent across families, model sizes, and calibration sources; outside it (e.g., for non-text modalities or architectures with substantially different routing or attention structure), they should be treated as hypotheses rather than established results.

Our comparisons against end-to-end fine-tuning and PEFT are necessarily restricted to model sizes at which these baselines are feasible to run, which is also the regime where local reconstruction is least needed. At larger scales, the case for local reconstruction relies on the consistency of its behavior across granularities and model sizes rather than on a direct head-to-head comparison with full-model adaptation. The claim that per-matrix reconstruction becomes competitive with block-wise reconstruction at the largest scales is supported by the largest models we evaluate and is consistent with the trend across smaller scales, but rests on a small number of points and should be extrapolated with caution.

Finally, our conclusions about reconstruction concern standard one-shot post-training pruning criteria with uniform sparsity and a fixed mask during adaptation, and assume the practitioner can afford a modest hyperparameter sweep over learning rate and number of epochs per submodel. \autoref{fig:qwen_lrs} suggests that the optimal learning rate is mostly stable across granularities, which mitigates but does not eliminate this cost.

\section{Impact Statement}
\label{app:impact_statement}

This work studies post-training compression of existing pretrained language models. We do not train new foundation models, do not introduce new capabilities or application domains, and do not release new datasets. Our contribution is methodological: we examine how to recover the quality of an already pruned model. The compressed models inherit all properties, including any biases, factual errors, or misuse potential, of the underlying pretrained checkpoints we start from, and our procedure neither amplifies nor mitigates these properties beyond what is already implied by reducing parameter count. The most direct broader effect of more effective post-pruning adaptation is a reduction in the inference cost and energy footprint of deploying large language models, which we view as a positive but unremarkable consequence of work in this area. We therefore do not identify ethical concerns or potential harms specific to this paper.

\section{Loss Function and Propagation Strategy}
\label{app:additional_experiments_loss_and_propagation}
\textbf{Propagation: How to define inputs and targets?}
Pruning a model locally involves splitting it into submodels and sequentially pruning each submodel. Before discussing the different propagation strategies, recall the pruning objective in \autoref{eq:layerwise_pruning_bg} and let us specify how the input activations $\mathbf{X}$ for the submodel $f$ we want to prune are computed. We denote by $\mathbf{X}:=f_0(\mathbf{X}_0;\theta_0)$ the output of the submodel $f_0$ consisting of all layers prior to $f$ with parameters $\theta_0$ based on input data $\mathbf{X}_0$. Further, we denote by $\mathbf{\hat{X}}:=f_0(\mathbf{X}_0;\hat{\theta}_0)$ the output of the submodel $f_0$ using the pruned weights $\hat{\theta}_0$. The different propagation strategies are defined by whether we use calibration inputs $\mathbf{\hat{X}}$ or $\mathbf{X}$, and similarly, whether to align the outputs with $f(\mathbf{X};\theta)$ or $f(\mathbf{\hat{X}};\theta)$. 
We outline three distinct strategies: 
\begin{enumerate}
    \item \gls{dp}: Both inputs and targets are sourced from the original dense model, i.e., the objective is $\min_{\hat{\theta}}\norm{f(\mathbf{X};\hat{\theta}) - f(\mathbf{X};\theta)}_F^2$. Here, the pruned submodel is reconstructed to best match the dense one, ignoring the error caused by pruning the prior submodels.
    \item \gls{sp}: Inputs come from the pruned model and targets are generated by further processing the sparse inputs through the dense model, i.e., the objective is $\min_{\hat{\theta}}\norm{f(\mathbf{\hat{X}};\hat{\theta}) - f(\mathbf{\hat{X}};\theta)}_F^2$. This approach is consistent with the EBFT method \citep{guo2024ebft}, where the reconstruction objective takes into account the error caused by pruning the prior submodels.
    \item \gls{mp}: Inputs come from the pruned model, while targets are sourced from the dense model, i.e., the objective is $\min_{\hat{\theta}}\norm{f(\mathbf{\hat{X}};\hat{\theta}) - f(\mathbf{X};\theta)}_F^2$. This approach tries to correct for the error caused by pruning the prior submodels by steering the activations to those of the dense model. \cite{Shin2024} refer to this strategy as \emph{global propagation}. 
\end{enumerate}

\textbf{Loss: How to measure reconstruction quality?}
The loss function measures the similarity between the outputs of the reconstructed submodel and the target activations. We consider two loss functions: \gls{mse}, the squared Euclidean distance averaged over batches, and \gls{cs}, a well-known measure of similarity between latent representations in NLP.

\autoref{tab:ppl_opt_1_3b}, \autoref{tab:ppl_opt_6_7b}, \autoref{tab:LLaMA-2_13b}, and \autoref{tab:LLaMA-3_8b} show the results of the different propagation strategies and loss functions for OPT-1.3B, OPT-6.7B, LLaMA-2-13B, and LLaMA-3-8B, respectively. Overall, we observe no systematic difference between the different propagation strategies and loss functions, both in terms of perplexity and zero-shot accuracy. Notably, for both OPT-1.3B and OPT-6.7B, finer granularities are not only on par with coarser granularities, but often better by a significant margin.

\begin{table}[H]
  \caption{Perplexity and zero-shot accuracy of OPT-6.7B pruned to different sparsity types with Wanda and reconstructed in different settings with 1024 calibration samples. The best result for each setting is underlined, the best result for each sparsity type is highlighted in bold. SP, MP, DP: sparse, mixed, and dense propagation. CS: cosine similarity. $\downarrow$: lower is better, $\uparrow$: higher is better. Base perplexity of OPT-6.7B dense: 10.86, pruned unstructured: 12.06, pruned 2:4: 16.05.\newline}
      \centering
      \tiny
      \begin{tabular}{cccccccccccccc}
      \toprule
        & & \multicolumn{6}{c}{Perplexity $\downarrow$} & \multicolumn{6}{c}{Zero-shot accuracy (in \%) $\uparrow$}\\
      \cmidrule(lr){3-8} \cmidrule(lr){9-14}
      Sparsity type & Block size & \multicolumn{2}{c}{SP} &  \multicolumn{2}{c}{MP} &  \multicolumn{2}{c}{DP} & \multicolumn{2}{c}{SP} &  \multicolumn{2}{c}{MP} &  \multicolumn{2}{c}{DP} \\
  \cmidrule(lr){3-4} \cmidrule(lr){5-6} \cmidrule(lr){7-8} \cmidrule(lr){9-10} \cmidrule(lr){11-12} \cmidrule(lr){13-14}
  & & MSE & CS & MSE & CS & MSE & CS & MSE & CS & MSE & CS & MSE & CS \\
  \midrule
  \multirow{4}{*}{\begin{tabular}[c]{@{}c@{}}50\%\\unstructured\end{tabular}} & $\sfrac{1}{2}$ & \underline{11.51} & \underline{\textbf{11.36}} & \underline{11.48} & \underline{11.37} & \underline{11.51} & \underline{\textbf{11.36}} & 49.94 & 50.08 & \underline{50.38} & \underline{50.47} & \underline{50.08} & 50.24 \\
    & 1 & 11.64 & 11.63 & 11.92 & 11.88 & 11.58 & 11.69 & 49.92 & \underline{50.62} & 50.03 & 50.20 & 49.82 & 50.52 \\
    & 2 & 11.59 & 11.54 & 11.75 & 11.91 & 11.61 & 11.53 & \underline{49.97} & 50.60 & 50.28 & 50.04 & 49.90 & \underline{\textbf{50.63}} \\
    & 8 & 11.85 & 11.89 & 11.89 & 11.95 & 11.84 & 11.90 & 49.84 & 49.76 & 50.03 & 49.90 & 49.76 & 49.78 \\
  \midrule
  \multirow{4}{*}{2:4} & $\sfrac{1}{2}$ & 13.41 & 13.01 & \underline{12.96} & \underline{\textbf{12.72}} & \underline{13.40} & 12.93 & 47.71 & 48.24 & \underline{48.92} & \underline{\textbf{48.94}} & 47.67 & 48.35 \\
    & 1 & 13.75 & 13.08 & 13.77 & 13.56 & 13.55 & 13.22 & 47.70 & \underline{48.77} & 48.31 & 48.75 & 47.82 & \underline{48.76} \\
    & 2 & \underline{13.30} & \underline{12.93} & 13.38 & 13.39 & 13.42 & \underline{12.85} & \underline{47.99} & 48.29 & 48.53 & 48.76 & \underline{47.94} & 48.66 \\
    & 8 & 13.49 & 13.45 & 13.56 & 13.57 & 13.54 & 13.45 & 47.93 & 48.14 & 48.07 & 48.28 & 47.86 & 48.09 \\
  \bottomrule
      \end{tabular}
      \label{tab:ppl_opt_6_7b}
  \end{table}

\begin{table}[H]
\caption{Perplexity and zero-shot accuracy of OPT-1.3B pruned to different sparsity types with Wanda and reconstructed in different settings with 1024 calibration samples. The best result for each setting is underlined, the best result for each sparsity type is highlighted in bold. SP, MP, DP: sparse, mixed, and dense propagation. CS: cosine similarity. $\downarrow$: lower is better, $\uparrow$: higher is better. Base perplexity of OPT-1.3B dense: 14.62, pruned unstructured: 18.31, pruned 2:4: 28.11.\newline}
    \centering
    \tiny
    \begin{tabular}{cccccccccccccc}
    \toprule
      & & \multicolumn{6}{c}{Perplexity $\downarrow$} & \multicolumn{6}{c}{Zero-shot accuracy (in \%) $\uparrow$}\\
    \cmidrule(lr){3-8} \cmidrule(lr){9-14}
    Sparsity type & Block size & \multicolumn{2}{c}{SP} &  \multicolumn{2}{c}{MP} &  \multicolumn{2}{c}{DP} & \multicolumn{2}{c}{SP} &  \multicolumn{2}{c}{MP} &  \multicolumn{2}{c}{DP} \\
\cmidrule(lr){3-4} \cmidrule(lr){5-6} \cmidrule(lr){7-8} \cmidrule(lr){9-10} \cmidrule(lr){11-12} \cmidrule(lr){13-14}
& & MSE & CS & MSE & CS & MSE & CS & MSE & CS & MSE & CS & MSE & CS \\
\midrule
\multirow{6}{*}{\begin{tabular}[c]{@{}c@{}}50\%\\unstructured\end{tabular}} & $\sfrac{1}{2}$ & \underline{15.60} & \underline{15.45} & \underline{15.47} & \underline{\textbf{15.43}} & \underline{15.60} & \underline{15.45} & \underline{44.99} & \underline{\textbf{45.16}} & \underline{44.63} & \underline{44.99} & \underline{44.68} & \underline{45.02} \\
  & 1 & 16.01 & 15.86 & 16.42 & 16.52 & 16.07 & 16.12 & 44.03 & 44.63 & 44.08 & 44.32 & 44.11 & 44.78 \\
  & 2 & 16.03 & 15.81 & 16.26 & 16.34 & 16.03 & 15.78 & 43.86 & 44.47 & 44.26 & 44.27 & 44.00 & 44.49 \\
  & 6 & 16.12 & 16.17 & 16.17 & 16.25 & 16.14 & 16.20 & 44.38 & 44.35 & 44.27 & 44.41 & 44.51 & 44.29 \\
  & 12 & 16.20 & 16.34 & 16.19 & 16.28 & 16.22 & 16.38 & 43.99 & 44.02 & 44.11 & 44.33 & 44.26 & 44.03 \\
  & 24 & 16.29 & 16.44 & 16.29 & 16.42 & 16.31 & 16.43 & 44.24 & 44.34 & 44.22 & 44.38 & 44.32 & 44.27 \\
\midrule
\multirow{6}{*}{2:4} & $\sfrac{1}{2}$ & \underline{18.49} & \underline{17.96} & \underline{\textbf{17.67}} & \underline{17.82} & \underline{18.52} & \underline{18.04} & \underline{43.39} & 43.39 & \underline{44.09} & \underline{\textbf{44.17}} & \underline{43.50} & 43.56 \\
  & 1 & 19.39 & 18.41 & 19.37 & 19.74 & 19.67 & 18.94 & 42.67 & \underline{43.53} & 42.97 & 43.27 & 42.66 & \underline{43.64} \\
  & 2 & 19.33 & 18.50 & 18.96 & 19.25 & 19.37 & 18.45 & 42.52 & 43.34 & 42.82 & 43.08 & 42.41 & 43.23 \\
  & 6 & 19.28 & 19.21 & 19.25 & 19.38 & 19.37 & 19.23 & 42.76 & 43.30 & 43.17 & 43.14 & 42.88 & 43.42 \\
  & 12 & 19.49 & 19.47 & 19.38 & 19.43 & 19.56 & 19.59 & 43.09 & 42.57 & 43.26 & 42.98 & 42.90 & 42.77 \\
  & 24 & 19.76 & 19.85 & 19.73 & 19.81 & 19.76 & 19.84 & 43.02 & 42.92 & 43.04 & 42.84 & 42.94 & 42.88 \\
\bottomrule
    \end{tabular}
    \label{tab:ppl_opt_1_3b}
\end{table}

\begin{table}[H]
\caption{Perplexity and zero-shot accuracy of LLaMA-2-13B (40 transformer blocks) pruned to different sparsity types with Wanda and reconstructed in different settings with 1024 calibration samples. The best result for each setting is underlined, the best result for each sparsity type is highlighted in bold. SP, MP, DP: sparse, mixed, and dense propagation. CS: cosine similarity. $\downarrow$: lower is better, $\uparrow$: higher is better. Base perplexity of LLaMA-2-13B dense: 4.57, pruned unstructured: 5.54, pruned 2:4: 8.39.\newline}
    \centering
    \tiny
    \begin{tabular}{cccccccccccccc}
    \toprule
      & & \multicolumn{6}{c}{Perplexity $\downarrow$} & \multicolumn{6}{c}{Zero-shot accuracy (in \%) $\uparrow$}\\
    \cmidrule(lr){3-8} \cmidrule(lr){9-14}
    Sparsity type & Block size & \multicolumn{2}{c}{SP} &  \multicolumn{2}{c}{MP} &  \multicolumn{2}{c}{DP} & \multicolumn{2}{c}{SP} &  \multicolumn{2}{c}{MP} &  \multicolumn{2}{c}{DP} \\
\cmidrule(lr){3-4} \cmidrule(lr){5-6} \cmidrule(lr){7-8} \cmidrule(lr){9-10} \cmidrule(lr){11-12} \cmidrule(lr){13-14}
& & MSE & CS & MSE & CS & MSE & CS & MSE & CS & MSE & CS & MSE & CS \\
\midrule
\multirow{4}{*}{unstructured} & $\sfrac{1}{2}$ & 5.33 & 5.33 & \underline{\textbf{5.26}} & 5.33 & 5.34 & 5.35 & 60.53 & 60.84 & 60.99 & 60.96 & 60.59 & 60.59 \\
  & 1 & 5.32 & 5.35 & \underline{\textbf{5.26}} & 5.34 & 5.33 & 5.37 & \underline{61.18} & \underline{61.03} & \underline{61.25} & \underline{\textbf{61.30}} & 61.02 & 60.75 \\
  & 2 & 5.32 & 5.36 & 5.27 & 5.33 & 5.33 & 5.37 & 60.93 & 60.45 & 60.91 & 60.84 & 60.62 & 60.61 \\
  & 10 & \underline{5.30} & \underline{5.31} & 5.29 & \underline{5.30} & \underline{5.31} & \underline{5.33} & 60.94 & 60.99 & 61.09 & 61.01 & \underline{61.06} & \underline{60.95} \\
\midrule
\multirow{4}{*}{2:4} & $\sfrac{1}{2}$ & 6.76 & 6.80 & \underline{\textbf{6.24}} & 6.59 & 6.71 & 6.80 & 56.16 & 56.15 & 58.66 & 58.30 & 56.34 & 56.10 \\
  & 1 & 6.68 & 6.72 & 6.32 & 6.63 & 6.67 & 6.79 & 56.42 & 56.47 & \underline{\textbf{58.75}} & \underline{58.52} & 56.61 & \underline{56.45} \\
  & 2 & 6.56 & 6.64 & 6.35 & 6.52 & 6.60 & 6.73 & 55.94 & 56.35 & 57.81 & 58.38 & \underline{56.93} & 56.33 \\
  & 10 & \underline{6.37} & \underline{6.37} & 6.33 & \underline{6.35} & \underline{6.39} & \underline{6.42} & \underline{56.64} & \underline{56.70} & 57.03 & 56.82 & 56.36 & 56.36 \\
\bottomrule
    \end{tabular}
    \label{tab:LLaMA-2_13b}
\end{table}

\begin{table}[H]
\caption{Perplexity and zero-shot accuracy of LLaMA-3-8B (32 transformer blocks) pruned to different sparsity types with Wanda and reconstructed in different settings with 1024 calibration samples. The best result for each setting is underlined, the best result for each sparsity type is highlighted in bold. SP, MP, DP: sparse, mixed, and dense propagation. CS: cosine similarity. $\downarrow$: lower is better, $\uparrow$: higher is better. Base perplexity of LLaMA-3-8B dense: 5.83, pruned unstructured: 8.96, pruned 2:4: 21.58.\newline}
    \centering
    \tiny
    \begin{tabular}{cccccccccccccc}
    \toprule
      & & \multicolumn{6}{c}{Perplexity $\downarrow$} & \multicolumn{6}{c}{Zero-shot accuracy (in \%) $\uparrow$}\\
    \cmidrule(lr){3-8} \cmidrule(lr){9-14}
    Sparsity type & Block size & \multicolumn{2}{c}{SP} &  \multicolumn{2}{c}{MP} &  \multicolumn{2}{c}{DP} & \multicolumn{2}{c}{SP} &  \multicolumn{2}{c}{MP} &  \multicolumn{2}{c}{DP} \\
\cmidrule(lr){3-4} \cmidrule(lr){5-6} \cmidrule(lr){7-8} \cmidrule(lr){9-10} \cmidrule(lr){11-12} \cmidrule(lr){13-14}
& & MSE & CS & MSE & CS & MSE & CS & MSE & CS & MSE & CS & MSE & CS \\
\midrule
\multirow{4}{*}{unstructured} & $\sfrac{1}{2}$ & 7.99 & 8.00 & 7.75 & 7.99 & 7.84 & 7.99 & 57.57 & 57.54 & \underline{58.81} & 57.44 & 58.80 & 57.64 \\
  & 1 & 7.76 & 8.00 & \underline{\textbf{7.70}} & 7.92 & 7.75 & 8.02 & \underline{58.55} & 57.77 & 58.08 & 58.45 & \underline{\textbf{58.84}} & 57.66 \\
  & 2 & \underline{7.75} & 7.98 & 7.72 & \underline{7.79} & \underline{7.74} & 8.02 & 58.31 & \underline{58.39} & 58.21 & \underline{58.67} & 58.29 & \underline{58.20} \\
  & 8 & 7.82 & \underline{7.89} & 7.86 & 7.83 & 7.83 & \underline{7.95} & 58.07 & 57.94 & 57.86 & 58.04 & 58.10 & 57.76 \\
  & 16 & 8.06 & 7.99 & 8.06 & 7.96 & 8.02 & 8.03 & 57.26 & 57.48 & 57.41 & 57.76 & 57.16 & 57.53 \\
\midrule
\multirow{4}{*}{2:4} & $\sfrac{1}{2}$ & 15.79 & 15.28 & \underline{\textbf{9.96}} & 11.96 & 11.36 & 12.78 & 46.91 & 47.58 & \underline{55.74} & 50.71 & 51.18 & 49.66 \\
  & 1 & 11.14 & 12.60 & 10.34 & 11.84 & 10.95 & 12.67 & \underline{52.48} & 50.99 & 54.55 & \underline{\textbf{55.77}} & \underline{52.65} & 50.55 \\
  & 2 & 10.74 & 11.90 & 10.34 & 11.18 & 10.56 & 12.04 & 52.00 & 51.24 & 52.78 & 53.33 & 51.51 & 50.16 \\
  & 8 & \underline{10.44} & \underline{10.59} & 10.57 & \underline{10.59} & \underline{10.52} & \underline{10.88} & 52.47 & \underline{52.51} & 52.92 & 53.44 & 52.40 & \underline{51.77} \\
  & 16 & 11.10 & 10.90 & 11.18 & 10.92 & 11.11 & 11.02 & 51.81 & 52.07 & 51.83 & 52.55 & 51.16 & 51.60 \\
\bottomrule
    \end{tabular}
    \label{tab:LLaMA-3_8b}
\end{table}

\section{Additional experiments with different calibration sets and benchmarks}
\label{app:additional_experiments_minipile}

\autoref{tab:minipile} shows perplexity and zero-shot accuracy of LLaMA-3-8B when pruned and reconstructed with a subset of MiniPile \citep{kaddour2023minipile} as calibration data. Comparing the results to \autoref{tab:LLaMA-3_8b}, we see that 1) the overall performance of the reconstructed models improves slightly when using MiniPile instead of C4, and 2) the relative performance differences between granularities, loss functions, and propagation strategies remain nearly identical.\\

\autoref{tab:qwen-2.5-32b-reasoning-wanda}, \autoref{tab:qwen-2.5-32b-reasoning-magnitude}, \autoref{tab:llama-3-8b-reasoning-wanda}, and \autoref{tab:llama-3-8b-reasoning-magnitude} show perplexity, zero-shot accuracy, code reasoning, and mathematical reasoning accuracy of Qwen-2.5-32B-Instruct and LLaMA-3-8B when pruned and reconstructed with a subset of C4, HumanEval \citep{chen2021codex}, or GSM8K \citep{cobbe2021gsm8k} as calibration data. Crucially, the central findings of our study (the gap between per-matrix and block-wise reconstruction, the free-lunch regime across granularities, and the diminishing importance of the pruning criterion at scale) all hold regardless of calibration dataset. Domain-specific calibration data can improve absolute performance on specialized tasks (e.g., using OpenCode for calibration increases the HumanEval coding score by 60\% over C4), but the relative behavior across granularities and pruning criteria remains unchanged.\\

\autoref{tab:llama-3-8B-cal-set-size} shows perplexity of LLaMA-3-8B when pruned and reconstructed with different block sizes and calibration set sizes from the C4 dataset. It shows that 1) increasing the granularity beyond block size $\sfrac{1}{2}$ does not improve the performance of the reconstructed model for a wide range of calibration set sizes, and 2) increasing the calibration set size beyond 256 samples does not improve the performance of the reconstructed model for per-matrix reconstruction.\\

\begin{table}[H]
\caption{Perplexity and zero-shot accuracy of LLaMA-3-8B (32 transformer blocks) pruned to different sparsity types with Wanda and reconstructed in different settings with 1024 calibration samples from the MiniPile dataset. The best result for each setting is underlined, the best result for each sparsity type is highlighted in bold. SP, MP, DP: sparse, mixed, and dense propagation. CS: cosine similarity. $\downarrow$: lower is better, $\uparrow$: higher is better. Base perplexity of LLaMA-3-8B dense: 5.83, pruned unstructured: 8.96, pruned 2:4: 21.58.\newline}
    \centering
    \tiny
    \begin{tabular}{cccccccccccccc}
    \toprule
      & & \multicolumn{6}{c}{Perplexity $\downarrow$} & \multicolumn{6}{c}{Zero-shot accuracy (in \%) $\uparrow$}\\
    \cmidrule(lr){3-8} \cmidrule(lr){9-14}
    Sparsity type & Block size & \multicolumn{2}{c}{SP} &  \multicolumn{2}{c}{MP} &  \multicolumn{2}{c}{DP} & \multicolumn{2}{c}{SP} &  \multicolumn{2}{c}{MP} &  \multicolumn{2}{c}{DP} \\
\cmidrule(lr){3-4} \cmidrule(lr){5-6} \cmidrule(lr){7-8} \cmidrule(lr){9-10} \cmidrule(lr){11-12} \cmidrule(lr){13-14}
& & MSE & CS & MSE & CS & MSE & CS & MSE & CS & MSE & CS & MSE & CS \\
\midrule
\multirow{5}{*}{unstructured} & $\sfrac{1}{2}$ & 7.85 & 7.98 & 7.68 & 7.89 & 7.84 & 7.98 & \underline{59.72} & 59.27 & \underline{60.40} & \underline{\textbf{61.04}} & 59.77 & 59.16 \\
  & 1 & 7.73 & 7.92 & 7.68 & 7.83 & 7.72 & 7.94 & 59.31 & \underline{59.55} & 60.04 & 60.23 & \underline{59.90} & \underline{59.40} \\
  & 2 & \underline{7.69} & 7.88 & \underline{\textbf{7.66}} & \underline{7.72} & \underline{7.69} & 7.92 & 59.16 & 59.39 & 59.52 & 60.25 & 59.17 & 59.39 \\
  & 8 & 7.71 & \underline{7.77} & 7.75 & 7.72 & 7.72 & \underline{7.83} & 58.81 & 58.87 & 59.00 & 59.15 & 58.68 & 58.59 \\
  & 16 & 8.03 & 7.97 & 7.90 & 7.82 & 7.91 & 7.87 & 57.62 & 57.86 & 58.08 & 58.33 & 57.86 & 58.16 \\
\midrule
\multirow{5}{*}{2:4} & $\sfrac{1}{2}$ & 11.48 & 12.77 & \underline{\textbf{9.65}} & 10.75 & 11.16 & 12.33 & 53.18 & 51.16 & \underline{56.50} & \underline{\textbf{56.66}} & 53.23 & 51.05 \\
  & 1 & 10.95 & 12.35 & 9.69 & 10.70 & 10.65 & 12.21 & \underline{54.28} & 52.10 & 56.31 & 56.60 & \underline{54.11} & 51.88 \\
  & 2 & 10.60 & 11.58 & 10.26 & 10.83 & 10.44 & 11.57 & 52.73 & 51.69 & 54.05 & 55.18 & 53.06 & 51.02 \\
  & 8 & \underline{10.15} & \underline{10.25} & 10.35 & \underline{10.29} & \underline{10.23} & \underline{10.55} & 53.32 & \underline{53.49} & 53.87 & 54.06 & 53.12 & \underline{52.42} \\
  & 16 & 11.82 & 10.46 & 10.82 & 10.52 & 10.77 & 10.61 & 50.24 & 52.00 & 52.80 & 53.48 & 52.27 & 52.29 \\
\bottomrule
    \end{tabular}
    \label{tab:minipile}
\end{table}

\begin{table}[H]
  \centering
  \caption{LLaMA-3-8B pruned to 2:4 sparsity with Wanda and reconstructed with different block sizes and calibration set sizes from the C4 dataset.\\ }
  \small
  \resizebox{\textwidth}{!}{%
  \begin{tabular}{@{}l *{10}{r}@{}}
    \toprule
    Cal.\ set size / Block size & 16 & 32 & 64 & 128 & 256 & 512 & 1024 & 2048 & 4096 & 8192 \\
    \midrule
    Per matrix & 20.85 & 19.99 & 19.37 & 18.86 & 18.81 & 18.77 & 18.80 & 18.79 & 18.80 & 18.76 \\
    1/2 & 15.43 & 14.07 & 13.69 & 13.07 & 12.16 & 11.51 & 11.24 & 11.18 & 10.98 & 10.78 \\
    1 & 14.78 & 13.70 & 13.33 & 12.77 & 12.06 & 11.49 & 11.29 & 11.25 & 10.95 & 10.82 \\
    8 & 15.11 & 13.84 & 13.36 & 12.81 & 12.09 & 11.54 & 11.26 & 11.20 & 10.92 & 10.84 \\
    \bottomrule
    \label{tab:llama-3-8B-cal-set-size}
  \end{tabular}%
  }
\end{table}

\begin{table}[H]
  \centering
  \caption{Qwen-2.5-32B-Instruct pruned to 2:4 sparsity with Wanda and reconstructed in different settings with 256 calibration samples.\\ }
  \small
  \begin{tabular}{@{}llrrrr@{}}
    \toprule
    Cal.\ Set & Block size & PPL & Acc. & HumanEval & GSM8K \\
    \midrule
    C4 & Per matrix & 14.42 & 66.84\% & 21.51\% & 65.19\% \\
    C4 & 1/2 & 6.39 & 69.04\% & 26.22\% & 72.56\% \\
    C4 & 1 & 6.40 & 69.10\% & 25.87\% & 73.31\% \\
    C4 & 8 & 6.37 & 69.08\% & 26.14\% & 73.26\% \\
    OpenCode & Per matrix & 15.26 & 61.85\% & 53.02\% & 71.13\% \\
    OpenCode & 1/2 & 8.11 & 65.80\% & 54.24\% & 73.96\% \\
    OpenCode & 1 & 8.08 & 65.88\% & 54.87\% & 74.14\% \\
    OpenCode & 8 & 8.02 & 65.77\% & 54.52\% & 74.09\% \\
    GSM8K & Per matrix & 14.85 & 61.62\% & 44.01\% & 75.52\% \\
    GSM8K & 1/2 & 8.72 & 64.18\% & 47.76\% & 79.77\% \\
    GSM8K & 1 & 8.68 & 64.24\% & 48.17\% & 79.98\% \\
    GSM8K & 8 & 8.71 & 64.25\% & 48.12\% & 79.93\% \\
    \bottomrule
    \label{tab:qwen-2.5-32b-reasoning-wanda}
  \end{tabular}
\end{table}

\begin{table}[H]
  \centering
  \caption{Qwen-2.5-32B-Instruct pruned to 2:4 sparsity with magnitude pruning and reconstructed in different settings with 256 calibration samples.\\ }
  \small
  \begin{tabular}{@{}llrrrr@{}}
    \toprule
    Cal.\ Set & Block size & PPL & Acc. & HumanEval & GSM8K \\
    \midrule
    C4 & Per matrix & 23.42 & 64.61\% & 15.10\% & 52.92\% \\
    C4 & 1/2 & 6.42 & 69.11\% & 26.07\% & 72.44\% \\
    C4 & 1 & 6.41 & 68.93\% & 26.03\% & 73.04\% \\
    C4 & 8 & 6.35 & 69.10\% & 25.96\% & 73.01\% \\
    OpenCode & Per matrix & 26.91 & 63.27\% & 33.02\% & 54.09\% \\
    OpenCode & 1/2 & 8.11 & 65.85\% & 54.21\% & 74.11\% \\
    OpenCode & 1 & 8.08 & 65.73\% & 54.43\% & 74.09\% \\
    OpenCode & 8 & 8.02 & 65.89\% & 54.22\% & 74.12\% \\
    GSM8K & Per matrix & 25.14 & 59.63\% & 27.14\% & 57.94\% \\
    GSM8K & 1/2 & 8.84 & 63.96\% & 47.82\% & 79.81\% \\
    GSM8K & 1 & 8.81 & 64.12\% & 48.03\% & 79.94\% \\
    GSM8K & 8 & 8.84 & 64.08\% & 47.96\% & 79.87\% \\
    \bottomrule
  \end{tabular}
  \label{tab:qwen-2.5-32b-reasoning-magnitude}
\end{table}

\begin{table}[H]
  \centering
  \caption{LLaMA-3-8B pruned to 2:4 sparsity with Wanda and reconstructed in different settings with 256 calibration samples.\\ }
  \small
  \begin{tabular}{@{}llrrrr@{}}
    \toprule
    Cal.\ Set & Block size & PPL & Acc. & HumanEval & GSM8K \\
    \midrule
    C4 & Per matrix & 18.81 & 48.26\% & 3.27\% & 5.22\% \\
    C4 & 1/2 & 10.02 & 55.52\% & 5.94\% & 10.38\% \\
    C4 & 1 & 10.19 & 55.69\% & 6.09\% & 10.06\% \\
    C4 & 8 & 10.85 & 54.40\% & 5.92\% & 9.89\% \\
    OpenCode & Per matrix & 19.93 & 46.71\% & 9.04\% & 8.15\% \\
    OpenCode & 1/2 & 14.08 & 51.88\% & 20.82\% & 15.21\% \\
    OpenCode & 1 & 13.89 & 52.71\% & 21.31\% & 15.38\% \\
    OpenCode & 8 & 14.83 & 51.15\% & 20.07\% & 14.46\% \\
    GSM8K & Per matrix & 19.39 & 45.04\% & 3.82\% & 12.26\% \\
    GSM8K & 1/2 & 15.89 & 48.47\% & 7.09\% & 21.85\% \\
    GSM8K & 1 & 15.56 & 50.61\% & 6.48\% & 23.49\% \\
    GSM8K & 8 & 15.28 & 49.27\% & 7.31\% & 22.44\% \\
    \bottomrule
  \end{tabular}
  \label{tab:llama-3-8b-reasoning-wanda}
\end{table}

\begin{table}[H]
  \centering
  \caption{LLaMA-3-8B pruned to 2:4 sparsity with magnitude pruning and reconstructed in different settings with 256 calibration samples.\\ }
  \small
  \begin{tabular}{@{}llrrrr@{}}
    \toprule
    Cal.\ Set & Block size & PPL & Acc. & HumanEval & GSM8K \\
    \midrule
    C4 & Per matrix & 45.81 & 44.60\% & 2.99\% & 3.14\% \\
    C4 & 1/2 & 12.73 & 53.39\% & 4.16\% & 5.90\% \\
    C4 & 1 & 12.97 & 53.76\% & 4.04\% & 5.77\% \\
    C4 & 8 & 13.14 & 52.43\% & 4.26\% & 5.91\% \\
    OpenCode & Per matrix & 68.55 & 39.14\% & 7.31\% & 7.08\% \\
    OpenCode & 1/2 & 26.29 & 46.09\% & 12.17\% & 10.12\% \\
    OpenCode & 1 & 24.75 & 45.04\% & 12.29\% & 10.67\% \\
    OpenCode & 8 & 25.38 & 44.71\% & 11.85\% & 10.26\% \\
    GSM8K & Per matrix & 56.31 & 41.12\% & 2.51\% & 8.25\% \\
    GSM8K & 1/2 & 18.67 & 48.36\% & 4.49\% & 18.43\% \\
    GSM8K & 1 & 20.17 & 48.06\% & 4.52\% & 18.14\% \\
    GSM8K & 8 & 19.31 & 46.92\% & 4.34\% & 17.29\% \\
    \bottomrule
  \end{tabular}
  \label{tab:llama-3-8b-reasoning-magnitude}
\end{table}

\section{Additional experiments with different sparsity types}
\label{app:structured_pruning}

In this section, we investigate how reconstruction affects different sparsity types. For structured sparsity, we compare Olica \citep{he2025olica} and FLAP \citep{an2024fluctuation} as the structured pruning method. \autoref{tab:llama-3-8b-structured} shows perplexity of LLaMA-3-8B when pruned to 20\% and 50\% structured sparsity with Olica and FLAP and reconstructed at different granularities. It shows that the trend of reconstruction shrinking the performance gap between pruning methods persists in the structured setting.

\autoref{tab:llama-3-8b-high-sparsity-wanda} and \autoref{tab:llama-3-8b-high-sparsity-magnitude} show perplexity of LLaMA-3-8B when pruned to different sparsities with Wanda and magnitude pruning and reconstructed at different block sizes. It shows that reconstruction improves model performance for a wide range of sparsities and the granularity trends observed in \autoref{sec:experiments} persist.\\

\begin{table}[H]
  \centering
  \caption{LLaMA-3-8B pruned to 20\% and 50\% structured sparsity and reconstructed at different granularities.\\ }
  \small
  \begin{tabular}{@{}lllrrrr@{}}
    \toprule
    Block size & Method & Sparsity & PPL no rec. & PPL rec. & Acc.\ no rec. & Acc rec. \\
    \midrule
    1 & FLAP & 20\% & 7.81 & 7.03 & 53.42\% & 57.68\% \\
    1 & Olica & 20\% & 7.25 & 6.98 & 54.11\% & 57.92\% \\
    1 & FLAP & 50\% & 46.25 & 17.51 & 42.87\% & 48.76\% \\
    1 & Olica & 50\% & 32.27 & 16.84 & 44.27\% & 49.14\% \\
    8 & FLAP & 20\% & 7.81 & 7.09 & 53.42\% & 57.52\% \\
    8 & Olica & 20\% & 7.25 & 7.04 & 54.11\% & 57.86\% \\
    8 & FLAP & 50\% & 46.25 & 17.18 & 42.87\% & 48.83\% \\
    8 & Olica & 50\% & 32.27 & 16.91 & 44.27\% & 49.08\% \\
    \bottomrule
  \end{tabular}
  \label{tab:llama-3-8b-structured}
\end{table}

\begin{table}[H]
  \centering
  \caption{LLaMA-3-8B pruned to different sparsities with Wanda and reconstructed at different block sizes.\\ }
  \small
  \begin{tabular}{@{}l *{6}{r}@{}}
    \toprule
    Sparsity / Block size & 0.4 & 0.5 & 0.6 & 0.7 & 0.8 & 0.9 \\
    \midrule
    No rec. & 7.37 & 9.12 & 18.11 & 87.93 & 691.64 & 18693.60 \\
    Per matrix & 7.28 & 8.78 & 15.32 & 21.91 & 66.22 & 3501.47 \\
    1/2 & 7.13 & 8.08 & 10.18 & 16.85 & 60.16 & 2490.68 \\
    1 & 7.19 & 8.14 & 11.14 & 16.65 & 59.70 & 1469.93 \\
    2 & 7.15 & 8.18 & 10.16 & 17.90 & 59.68 & 2866.76 \\
    8 & 7.20 & 8.26 & 10.86 & 19.20 & 60.35 & 1272.12 \\
    \bottomrule
  \end{tabular}
  \label{tab:llama-3-8b-high-sparsity-wanda}
\end{table}

\begin{table}[H]
  \centering
  \caption{LLaMA-3-8B pruned to different sparsities with magnitude pruning and reconstructed at different block sizes.\\ }
  \small
  \begin{tabular}{@{}l *{6}{r}@{}}
    \toprule
    Sparsity / Block size & 0.4 & 0.5 & 0.6 & 0.7 & 0.8 & 0.9 \\
    \midrule
    No rec. & 32.25 & 96.12 & 352.48 & 847.28 & $2.64\times 10^{5}$ & $5.82\times 10^{6}$ \\
    Per matrix & 12.18 & 19.64 & 31.33 & 54.52 & 76623.15 & $3.57\times 10^{5}$ \\
    1/2 & 7.39 & 8.71 & 11.97 & 22.98 & 46267.20 & 37468.30 \\
    1 & 8.62 & 9.81 & 12.57 & 21.34 & 21348.80 & 13152.60 \\
    2 & 7.53 & 8.81 & 11.95 & 21.21 & 44843.70 & $2.34\times 10^{5}$ \\
    8 & 7.65 & 9.41 & 12.59 & 24.56 & 123.04 & 21685.00 \\
    \bottomrule
  \end{tabular}
  \label{tab:llama-3-8b-high-sparsity-magnitude}
\end{table}

\section{Additional Experiments on Granularity}
\label{app:additional_experiments_per_matrix}
Together with \autoref{tab:ppl_opt_1_3b}, \autoref{tab:ppl_opt_6_7b}, \autoref{tab:LLaMA-2_13b}, and \autoref{tab:LLaMA-3_8b} in \autoref{app:additional_experiments_loss_and_propagation}, we present additional experiments ablating the effect of granularity on the final model quality. Additionally, \autoref{tab:ppl_opt_1_3b_no_ln}, \autoref{tab:ppl_opt_6_7b_no_ln} compare per-matrix reconstruction with multiple coarser granularities under different loss functions and propagation strategies. One clear trend is that per-matrix reconstruction heavily underperforms in the mixed propagation setting. Hence, \autoref{tab:all_llama2_per_matrix} and \autoref{tab:all_qwen_per_matrix} compare per-matrix reconstruction using sparse propagation with block-wise reconstruction using mixed propagation (our default setting).

Overall, the trend observed in \autoref{sec:experiments} persists. Apart from per-matrix reconstruction, the final model quality is either largely insensitive to granularity or even slightly worse for coarser granularities and smaller models.

\begin{table}[H]
  \caption{Perplexity and average zero-shot accuracy of different LLaMA-2 models when pruned and reconstructed in different settings. The calibration set size is 256 and the pruning method is Wanda. Sparsities 50\% and 60\% are unstructured. The best result for each model, reconstruction mode, prune method combination is highlighted in bold. $\downarrow$: lower is better, $\uparrow$: higher is better.\newline}
  \label{tab:all_llama2_per_matrix}
  \setlength{\tabcolsep}{5pt}
  \tiny
      \centering
      \begin{tabular}{cccccccccccccc}
  \toprule
  \textbf{Perplexity} $\downarrow$ & & \multicolumn{3}{c}{\textbf{7B} (4.86)} & \multicolumn{3}{c}{\textbf{13B} (4.42)} & \multicolumn{3}{c}{\textbf{70B} (-)}\\
  \cmidrule(lr){3-5} \cmidrule(lr){6-8} \cmidrule(lr){9-11}
  Method & Sparsity & No Rec. & Per-matrix & Block size 1 & No Rec. & Per-matrix & Block size 1 & No Rec. & Per-matrix & Block size 1\\
  \midrule
  Magnitude & 50\% & 12.77 & 6.70 & \textbf{6.12} & 6.05 & 5.52 & \textbf{5.28} & 4.71 & 3.85 & \textbf{3.69}\\
  SparseGPT & 50\% & 6.13 & 6.11 & \textbf{5.92} & 5.48 & 5.37 & \textbf{5.25} & 3.81 & 3.79 & \textbf{3.71}\\
  Wanda & 50\% & 6.05 & 6.02 & \textbf{5.86} & 5.37 & 5.29 & \textbf{5.18} & 3.69 & 3.66 & \textbf{3.58}\\
  \midrule
  Magnitude & 60\% & 1370.12 & 14.44 & \textbf{8.08} & 9.58 & 8.41 & \textbf{6.43} & 7.93 & 5.16 & \textbf{4.40}\\
  SparseGPT & 60\% & 8.37 & 8.24 & \textbf{7.11} & 7.14 & 7.10 & \textbf{6.31} & 4.86 & 4.81 & \textbf{4.70}\\
  Wanda & 60\% & 8.97 & 8.79 & \textbf{7.08} & 7.37 & 7.22 & \textbf{6.21} & 4.75 & 4.72 & \textbf{4.52}\\
  \midrule
  Magnitude & 2:4 & 36.09 & 11.33 & \textbf{10.06} & 7.37 & 7.30 & \textbf{7.04} & 6.23 & 6.21 & \textbf{4.60}\\
  SparseGPT & 2:4 & 9.14 & 9.12 & \textbf{7.01} & 7.62 & 7.58 & \textbf{6.39} & 5.22 & 5.18 & \textbf{4.79}\\
  Wanda & 2:4 & 9.90 & 9.86 & \textbf{6.98} & 7.71 & 7.59 & \textbf{6.22} & 5.02 & 5.01 & \textbf{4.56}\\
  \midrule
  Magnitude & 4:8 & 12.97 & 8.21 & \textbf{6.63} & 6.41 & 6.25 & \textbf{5.87} & 5.34 & 5.28 & \textbf{4.12}\\
  SparseGPT & 4:8 & 7.10 & 7.02 & \textbf{6.47} & 6.38 & 6.27 & \textbf{5.77} & 4.51 & 4.45 & \textbf{4.30}\\
  Wanda & 4:8 & 7.28 & 7.17 & \textbf{6.36} & 6.25 & 6.21 & \textbf{5.71} & 4.24 & 4.19 & \textbf{4.08}\\
  \midrule
  \textbf{Accuracy} $\uparrow$ & & \multicolumn{3}{c}{\textbf{7B} (59.70\%)} & \multicolumn{3}{c}{\textbf{13B} (63.04\%)} & \multicolumn{3}{c}{\textbf{70B} (-)}\\
  \cmidrule(lr){3-5} \cmidrule(lr){6-8} \cmidrule(lr){9-11}
  Method & Sparsity & No Rec. & Per-matrix & Block size 1 & No Rec. & Per-matrix & Block size 1 & No Rec. & Per-matrix & Block size 1\\
  \midrule
  Magnitude & 50\% & 51.12\% & 54.90\% & \textbf{55.95\%} & 52.82\% & 59.45\% & \textbf{60.42\%} & 59.81\% & 65.63\% & \textbf{65.88\%}\\
  SparseGPT & 50\% & 56.44\% & 56.48\% & \textbf{56.71\%} & 59.82\% & 60.26\% & \textbf{60.91\%} & 66.19\% & 66.08\% & \textbf{66.48\%}\\
  Wanda & 50\% & 56.03\% & 56.08\% & \textbf{56.14\%} & 60.34\% & 60.40\% & \textbf{60.98\%} & 66.18\% & 66.20\% & \textbf{66.51\%}\\
  \midrule
  Magnitude & 60\% & 40.11\% & 46.88\% & \textbf{51.76\%} & 43.91\% & 44.74\% & \textbf{55.78\%} & 55.26\% & 63.35\% & \textbf{63.88\%}\\
  SparseGPT & 60\% & 51.78\% & 51.92\% & \textbf{52.52\%} & 56.46\% & 56.92\% & \textbf{58.03\%} & 64.87\% & 64.89\% & \textbf{65.12\%}\\
  Wanda & 60\% & 49.75\% & 50.28\% & \textbf{53.25\%} & 56.20\% & 56.43\% & \textbf{58.02\%} & 63.75\% & 63.81\% & \textbf{63.99\%}\\
  \midrule
  Magnitude & 2:4 & 47.47\% & 48.90\% & \textbf{49.02\%} & 49.85\% & 52.02\% & \textbf{53.46\%} & 58.87\% & 59.02\% & \textbf{63.12\%}\\
  SparseGPT & 2:4 & 51.57\% & 50.92\% & \textbf{52.97\%} & 55.92\% & 55.96\% & \textbf{57.86\%} & 62.47\% & 62.56\% & \textbf{63.67\%}\\
  Wanda & 2:4 & 48.61\% & 49.17\% & \textbf{52.36\%} & 53.67\% & 54.15\% & \textbf{56.49\%} & 62.86\% & 63.14\% & \textbf{63.84\%}\\
  \midrule
  Magnitude & 4:8 & 50.68\% & 52.42\% & \textbf{53.73\%} & 52.81\% & 55.96\% & \textbf{57.31\%} & 59.32\% & 59.39\% & \textbf{64.72\%}\\
  SparseGPT & 4:8 & 53.91\% & 54.22\% & \textbf{54.33\%} & 58.64\% & 59.49\% & \textbf{59.67\%} & 64.75\% & 64.86\% & \textbf{65.16\%}\\
  Wanda & 4:8 & 52.54\% & 53.03\% & \textbf{54.67\%} & 58.23\% & 58.30\% & \textbf{58.53\%} & 65.06\% & 65.11\% & \textbf{65.14\%}\\
  \bottomrule
      \end{tabular}
\end{table}

\begin{table}[H]
  \caption{Perplexity and zero-shot accuracy of OPT-1.3B pruned to different sparsity types with Wanda and reconstructed in different settings with 1024 calibration samples. The best result for each setting is underlined, the best result for each sparsity type is highlighted in bold. SP, MP, DP: sparse, mixed, and dense propagation. CS: cosine similarity. $\downarrow$: lower is better, $\uparrow$: higher is better. Base perplexity of OPT-1.3B dense: 14.62, pruned unstructured: 18.31, pruned 2:4: 28.11.\newline}
  \vspace{10pt}
      \centering
      \tiny
      \begin{tabular}{cccccccccccccc}
      \toprule
        & & \multicolumn{6}{c}{Perplexity $\downarrow$} & \multicolumn{6}{c}{Zero-shot accuracy (in \%) $\uparrow$}\\
      \cmidrule(lr){3-8} \cmidrule(lr){9-14}
      Sparsity type & Block size & \multicolumn{2}{c}{SP} &  \multicolumn{2}{c}{MP} &  \multicolumn{2}{c}{DP} & \multicolumn{2}{c}{SP} &  \multicolumn{2}{c}{MP} &  \multicolumn{2}{c}{DP} \\
  \cmidrule(lr){3-4} \cmidrule(lr){5-6} \cmidrule(lr){7-8} \cmidrule(lr){9-10} \cmidrule(lr){11-12} \cmidrule(lr){13-14}
  & & MSE & CS & MSE & CS & MSE & CS & MSE & CS & MSE & CS & MSE & CS \\
  \midrule
  \multirow{5}{*}{\begin{tabular}[c]{@{}c@{}}50\%\\unstructured\end{tabular}} & Per-matrix & 17.31 & 17.88 & 21.74 & 22.01 & 18.41 & 18.32 & 44.13 & 43.70 & 42.54 & 42.47 & 43.42 & 43.56 \\
    & $\sfrac{1}{2}$ & \underline{15.53} & \underline{15.46} & \underline{15.47} & \underline{\textbf{15.44}} & \underline{15.59} & \underline{15.46} & \underline{44.80} & \underline{\textbf{45.32}} & \underline{44.86} & \underline{45.03} & \underline{44.94} & \underline{45.29} \\
    & 1 & 16.02 & 15.91 & 16.42 & 16.54 & 16.09 & 16.13 & 44.16 & 44.80 & 44.05 & 44.30 & 44.27 & 44.91 \\
    & 2 & 16.03 & 15.82 & 16.25 & 16.35 & 16.04 & 15.80 & 43.94 & 44.40 & 44.27 & 44.28 & 44.00 & 44.55 \\
    & 6 & 16.12 & 16.17 & 16.17 & 16.25 & 16.14 & 16.20 & 44.47 & 44.36 & 44.30 & 44.19 & 44.45 & 44.34 \\
  \midrule
  \multirow{5}{*}{2:4} & Per-matrix & 24.48 & 26.25 & 42.62 & 46.41 & 27.96 & 27.69 & 41.84 & 41.08 & 40.35 & 40.86 & 41.44 & 41.84 \\
    & $\sfrac{1}{2}$ & \underline{18.50} & \underline{18.02} & \underline{\textbf{17.77}} & \underline{17.85} & \underline{18.59} & \underline{18.09} & \underline{43.34} & 43.42 & \underline{44.04} & \underline{\textbf{44.26}} & \underline{43.21} & 43.57 \\
    & 1 & 19.72 & 18.42 & 19.40 & 19.78 & 19.86 & 18.92 & 42.73 & \underline{43.52} & 43.54 & 43.18 & 42.76 & \underline{43.63} \\
    & 2 & 19.15 & 18.47 & 18.99 & 19.22 & 19.38 & 18.41 & 42.59 & 43.41 & 42.90 & 43.15 & 42.38 & 43.16 \\
    & 6 & 19.33 & 19.24 & 19.48 & 19.71 & 19.41 & 19.26 & 42.75 & 43.09 & 42.63 & 42.66 & 42.89 & 43.41 \\
  \bottomrule
      \end{tabular}
      \label{tab:ppl_opt_1_3b_no_ln}
  \end{table}

  \begin{table}[H]
    \caption{Perplexity and average zero-shot accuracy of different Qwen-2.5-Instruct models when pruned and reconstructed in different settings. The calibration set size is 256 and the pruning method is Wanda. Sparsities 50\% and 60\% are unstructured. The best result for each model, reconstruction mode, prune method combination is highlighted in bold. $\downarrow$: lower is better, $\uparrow$: higher is better.\newline}
    \label{tab:all_qwen_per_matrix}
    \setlength{\tabcolsep}{1.4pt}
    \tiny
        \centering
        \begin{tabular}{cccccccccccccc}
    \toprule
    \textbf{Perplexity} $\downarrow$ & & \multicolumn{3}{c}{\textbf{7B} (5.16)} & \multicolumn{3}{c}{\textbf{14B} (2.39)} & \multicolumn{3}{c}{\textbf{32B} (2.19)} & \multicolumn{3}{c}{\textbf{72B} (1.88)}\\
    \cmidrule(lr){3-5} \cmidrule(lr){6-8} \cmidrule(lr){9-11} \cmidrule(lr){12-14}
    Method & Sparsity & No Rec. & Per-matrix & Block size 1 & No Rec. & Per-matrix & Block size 1 & No Rec. & Per-matrix & Block size 1 & No Rec. & Per-matrix & Block size 1\\
    \midrule
    Magnitude & 50\%  & 568.93 & 9.92 & \textbf{7.94} & 18.27 & 8.12 & \textbf{7.71} & 13.22 & 4.90 & \textbf{4.36} & 444.81 & 4.34 & \textbf{4.28} \\
    SparseGPT & 50\%  & 7.59 & 7.47 & \textbf{7.41} & 6.68 & 6.59 & \textbf{6.55} & 4.84 & 4.81 & \textbf{4.69} & 4.20 & 4.17 & \textbf{4.03} \\
    Wanda & 50\%  & 7.67 & 7.52 & \textbf{7.40} & 6.42 & 6.38 & \textbf{6.12} & 4.43 & 4.42 & \textbf{4.25} & 3.70 & \textbf{3.64} & 3.65 \\
    \midrule
    Magnitude & 60\%  & 25551.15 & 31.35 & \textbf{12.35} & 60343.74 & 967.78 & \textbf{10.75} & 49.04 & 9.00 & \textbf{6.31} & 1071.23 & 7.80 & \textbf{6.02} \\
    SparseGPT & 60\%  & 13.33 & 11.97 & \textbf{9.60} & 117.86 & 50.59 & \textbf{9.69} & 6.62 & 6.60 & \textbf{6.47} & 6.38 & 6.29 & \textbf{6.22} \\
    Wanda & 60\%  & 26.51 & 15.64 & \textbf{9.29} & 170.16 & 76.10 & \textbf{9.23} & 62.11 & 7.02 & \textbf{6.28} & 5.78 & \textbf{5.68} & 5.70 \\
    \midrule
    Magnitude & 2:4  & 5974.84 & 15.31 & \textbf{10.18} & 1424.70 & 387.53 & \textbf{12.11} & 38.34 & 67.16 & \textbf{6.31} & 226.30 & 6.88 & \textbf{5.93} \\
    SparseGPT & 2:4  & 13.02 & 12.77 & \textbf{9.69} & 198.16 & 89.20 & \textbf{10.87} & 7.11 & 7.04 & \textbf{6.71} & 6.83 & 6.74 & \textbf{6.39} \\
    Wanda & 2:4  & 15.98 & 15.19 & \textbf{9.60} & 242.79 & 58.58 & \textbf{12.85} & 37.52 & 13.89 & \textbf{6.50} & 5.99 & 5.96 & \textbf{5.86} \\
    \midrule
    Magnitude & 4:8  & 4138.65 & 53.75 & \textbf{8.81} & 87.93 & 68.21 & \textbf{7.85} & 24.51 & 6.04 & \textbf{5.46} & 195.85 & 5.72 & \textbf{4.75} \\
    SparseGPT & 4:8  & 8.95 & 8.76 & \textbf{8.21} & 56.11 & 35.51 & \textbf{7.12} & 6.10 & \textbf{5.86} & 5.88 & 5.42 & 5.40 & \textbf{5.33} \\
    Wanda & 4:8  & 9.43 & 9.32 & \textbf{8.20} & 72.05 & 28.38 & \textbf{7.07} & 5.79 & 5.79 & \textbf{5.47} & 4.91 & 4.87 & \textbf{4.82} \\
    \midrule
    \textbf{Accuracy} $\uparrow$ & & \multicolumn{3}{c}{\textbf{7B} (67.53\%)} & \multicolumn{3}{c}{\textbf{14B} (71.05\%)} & \multicolumn{3}{c}{\textbf{32B} (69.66\%)} & \multicolumn{3}{c}{\textbf{72B} (72.32\%)}\\
    \cmidrule(lr){3-5} \cmidrule(lr){6-8} \cmidrule(lr){9-11} \cmidrule(lr){12-14}
    Method & Sparsity & No Rec. & Per-matrix & Block size 1 & No Rec. & Per-matrix & Block size 1 & No Rec. & Per-matrix & Block size 1 & No Rec. & Per-matrix & Block size 1\\
    \midrule
    Magnitude & 50\%  & 46.87\% & 63.30\% & \textbf{64.50\%} & 61.56\% & 67.41\% & \textbf{67.55\%} & 61.94\% & 67.52\% & \textbf{68.28\%} & 51.27\% & 70.86\% & \textbf{71.16\%} \\
    SparseGPT & 50\%  & 64.42\% & 65.22\% & \textbf{65.31\%} & 68.75\% & 68.89\% & \textbf{68.94\%} & 68.42\% & 68.42\% & \textbf{69.06\%} & 71.25\% & 71.42\% & \textbf{71.76\%} \\
    Wanda & 50\%  & 62.95\% & 64.01\% & \textbf{64.88\%} & 68.36\% & 68.52\% & \textbf{68.76\%} & 68.03\% & 68.42\% & \textbf{68.72\%} & 70.94\% & 71.53\% & \textbf{71.72\%} \\
    \midrule
    Magnitude & 60\%  & 35.41\% & 57.85\% & \textbf{60.58\%} & 50.38\% & 63.34\% & \textbf{64.62\%} & 55.24\% & 65.77\% & \textbf{67.05\%} & 46.27\% & 68.15\% & \textbf{69.31\%} \\
    SparseGPT & 60\%  & 60.23\% & 61.53\% & \textbf{62.40\%} & 65.48\% & 65.37\% & \textbf{65.74\%} & 66.14\% & 66.56\% & \textbf{66.98\%} & 69.75\% & 70.02\% & \textbf{70.22\%} \\
    Wanda & 60\%  & 60.67\% & 61.77\% & \textbf{62.00\%} & 63.87\% & 64.66\% & \textbf{65.30\%} & 66.09\% & 66.29\% & \textbf{67.30\%} & 68.99\% & 69.78\% & \textbf{70.18\%} \\
    \midrule
    Magnitude & 2:4  & 41.23\% & 60.13\% & \textbf{61.52\%} & 53.83\% & 61.22\% & \textbf{65.02\%} & 58.55\% & 65.93\% & \textbf{67.32\%} & 51.07\% & 68.21\% & \textbf{69.03\%} \\
    SparseGPT & 2:4  & 60.81\% & 60.97\% & \textbf{61.63\%} & 64.76\% & 64.87\% & \textbf{65.63\%} & 66.33\% & 66.38\% & \textbf{67.26\%} & 69.20\% & 69.39\% & \textbf{69.44\%} \\
    Wanda & 2:4  & 59.18\% & 60.07\% & \textbf{61.52\%} & 63.75\% & 64.49\% & \textbf{65.68\%} & 66.41\% & 67.10\% & \textbf{67.39\%} & 69.31\% & \textbf{69.99\%} & 69.94\% \\
    \midrule
    Magnitude & 4:8  & 44.65\% & 55.16\% & \textbf{63.34\%} & 58.70\% & 60.18\% & \textbf{66.51\%} & 59.47\% & 67.17\% & \textbf{68.13\%} & 52.22\% & 69.38\% & \textbf{69.61\%} \\
    SparseGPT & 4:8  & 62.58\% & 63.21\% & \textbf{63.80\%} & 67.14\% & 67.16\% & \textbf{67.76\%} & 67.35\% & 67.78\% & \textbf{68.06\%} & 69.96\% & 70.16\% & \textbf{70.92\%} \\
    Wanda & 4:8  & 61.61\% & 62.67\% & \textbf{62.96\%} & 65.59\% & 66.32\% & \textbf{66.93\%} & 67.54\% & 68.29\% & \textbf{68.53\%} & 69.82\% & 70.41\% & \textbf{70.54\%} \\
    \bottomrule
        \end{tabular}
  \end{table}

  \begin{table}[H]
    \caption{Perplexity and zero-shot accuracy of OPT-6.7B pruned to different sparsity types with Wanda and reconstructed at various granularities and 1024 calibration samples. Layer norm parameters are fixed for comparability with per-matrix reconstruction. The best result for each setting is underlined, the best result for each sparsity type is highlighted in bold. $\downarrow$: lower is better, $\uparrow$: higher is better. Dense baseline: 10.86 perplexity, pruned unstructured: 12.06, pruned 2:4: 16.05.\newline}
        \centering
        \tiny
        \begin{tabular}{cccccccccccccc}
        \toprule
          & & \multicolumn{6}{c}{Perplexity $\downarrow$} & \multicolumn{6}{c}{Zero-shot accuracy (in \%) $\uparrow$}\\
        \cmidrule(lr){3-8} \cmidrule(lr){9-14}
        Sparsity type & Block size & \multicolumn{2}{c}{SP} &  \multicolumn{2}{c}{MP} &  \multicolumn{2}{c}{DP} & \multicolumn{2}{c}{SP} &  \multicolumn{2}{c}{MP} &  \multicolumn{2}{c}{DP} \\
    \cmidrule(lr){3-4} \cmidrule(lr){5-6} \cmidrule(lr){7-8} \cmidrule(lr){9-10} \cmidrule(lr){11-12} \cmidrule(lr){13-14}
    & & MSE & CS & MSE & CS & MSE & CS & MSE & CS & MSE & CS & MSE & CS \\
    \midrule
    \multirow{5}{*}{unstructured} & Per-matrix & 11.55 & 11.72 & 98.15 & 49.72 & 11.96 & 11.99 & 49.39 & 49.48 & 43.64 & 44.95 & 48.78 & 48.98 \\
      & $\sfrac{1}{2}$ & \underline{11.49} & \underline{\textbf{11.27}} & \underline{11.66} & \underline{11.34} & \underline{11.51} & \underline{11.40} & 49.99 & \underline{\textbf{50.62}} & \underline{50.60} & \underline{50.49} & 49.86 & 50.14 \\
      & 1 & 11.65 & 11.59 & 11.95 & 11.90 & 11.56 & 11.71 & 49.85 & 50.54 & 49.97 & 50.27 & 49.86 & 50.61 \\
      & 2 & 11.61 & 11.56 & 11.78 & 11.84 & 11.60 & 11.57 & \underline{50.02} & 50.52 & 50.23 & 50.18 & \underline{49.92} & \underline{\textbf{50.62}} \\
      & 8 & 11.84 & 11.90 & 11.86 & 11.91 & 11.88 & 11.93 & 49.73 & 49.75 & 50.02 & 49.89 & 49.84 & 49.68 \\
    \midrule
    \multirow{5}{*}{2:4} & Per-matrix & 14.25 & 15.82 & 293.26 & 104.03 & 16.06 & 15.52 & 47.02 & 46.94 & 38.42 & 40.62 & 45.53 & 45.99 \\
      & $\sfrac{1}{2}$ & 13.39 & 12.98 & \underline{12.97} & \underline{\textbf{12.72}} & 13.45 & 12.97 & 47.87 & 48.18 & \underline{\textbf{48.90}} & \underline{48.80} & 47.84 & 48.28 \\
      & 1 & \underline{13.29} & 13.04 & 13.85 & 13.58 & 13.59 & 13.21 & 47.91 & \underline{48.61} & 48.31 & 48.59 & 47.84 & \underline{48.76} \\
      & 2 & 13.43 & \underline{12.93} & 13.46 & 13.40 & \underline{13.41} & \underline{12.90} & 47.83 & 48.47 & 48.48 & 48.78 & \underline{48.03} & 48.75 \\
      & 8 & 13.47 & 13.51 & 13.52 & 13.55 & 13.64 & 13.24 & \underline{47.92} & 47.98 & 48.35 & 48.31 & 47.82 & 48.25 \\
    \bottomrule
        \end{tabular}
        \label{tab:ppl_opt_6_7b_no_ln}
  \end{table}

\section{Hyperparameters}
\label{app:hyperparameters}

Here, we provide the hyperparameter grids used for grid search for local reconstruction, full retraining, and MaskLoRA fine-tuning.

\begin{table}[h]
    \caption{Hyperparameter grids used for grid search for local reconstruction, full retraining, and MaskLoRA fine-tuning.\newline}
    \tiny
    \vspace{10pt}
        \centering
        \begin{tabular}{ccccccccccc}
        \toprule
        Method & Learning rate & Warmup ratio & Scheduler & Number of epochs & Optimizer \\
        \midrule
        Local reconstruction & \{1e-6, 5e-6, 1e-5, 5e-5, 1e-4\} & 0.1 & Linear & $\{1,5,10\}$ & AdamW \\
        MaskLoRA fine-tuning & \{5e-7, 1e-6, 5e-6, 1e-5, 5e-5, 1e-4, 5e-4, 1e-3\} & 0.1 & Linear & $\{1,5,10\}$ & AdamW \\
        Full retraining & \{5e-7, 1e-6, 5e-6, 1e-5, 5e-5, 1e-4, 5e-4, 1e-3\} & 0.1 & Linear & $\{1,5,10\}$ & AdamW \\
        \midrule
        Method & Batch size & Gradient accumulation & LoRA rank & LoRA alpha & Max. gradient norm \\
        \midrule
        Local reconstruction & 2 & 1 & - & - & - \\
        MaskLoRA fine-tuning & 1 & 2 & 16 & 32 & \{0.5,1.0,1.5\} \\
        Full retraining & 1 & 2 & - & - & \{0.5,1.0,1.5\} \\
        \bottomrule
        \end{tabular}
        \label{tab:hyperparameters}
\end{table}

\end{document}

%% file: granularity_figure2.tex
\def\textoffset{-0.25}
\def\blockheight{2}
\def\matrixoffset{0.1}
\def\boxoffset{0.15}

\begin{tikzpicture}[x=0.8cm,y=0.8cm, font=\scriptsize,
  ]

  \draw[draw=plotcolor5, fill=plotcolor5!20, thick, rounded corners, dashed] (-0.5-3*\boxoffset,-0.75-2*\boxoffset) rectangle (7+8.5+3*\boxoffset, \blockheight+3*\boxoffset-6*\textoffset);
  \node[] at (7.5, \blockheight+2.5*\boxoffset-5*\textoffset) {\color{plotcolor5}Block size 2};

  \def\blockoffset{0.0}

  \draw[draw=plotcolor4, fill=plotcolor4!20, thick, rounded corners, dashed] (-0.5+\blockoffset-2*\boxoffset,-0.75-\boxoffset) rectangle (7+\blockoffset+2*\boxoffset, \blockheight+2*\boxoffset-4*\textoffset);
  \node[] at (3.25+\blockoffset, \blockheight+1.5*\boxoffset-3*\textoffset) {\color{plotcolor4}Block size 1};
  \draw[draw=plotcolor2, fill=plotcolor2!20, thick, rounded corners, dashed] (-0.5+\blockoffset-\boxoffset,-0.75) rectangle (3+\blockoffset+\boxoffset, \blockheight+\boxoffset-2*\textoffset);
  \node[] at (1.25+\blockoffset, \blockheight+0.5*\boxoffset-\textoffset) {\color{plotcolor2}Block size $\sfrac{1}{2}$};
  \draw[draw=plotcolor2, fill=plotcolor2!20, thick, rounded corners, dashed] (3.5+\blockoffset-\boxoffset,-0.75) rectangle (7+\blockoffset+\boxoffset, \blockheight+\boxoffset-2*\textoffset);
  \node[] at (5.25+\blockoffset, \blockheight+0.5*\boxoffset-\textoffset) {\color{plotcolor2}Block size $\sfrac{1}{2}$};

  \draw[draw=black, fill=white, thick, rounded corners] (-0.25+\blockoffset,0) rectangle (2+\blockoffset, \blockheight);
  \node[] at (0.875+\blockoffset,\blockheight+\textoffset) {Attention};

  \def\matrixwidth{0.5625}
  \draw[draw=black, thick, rounded corners] (-0.25+\blockoffset+\matrixoffset,0+\matrixoffset) rectangle (-0.25+\matrixwidth+\blockoffset-\matrixoffset*0.25, \blockheight+2*\textoffset);
  \draw[draw=black, thick, rounded corners] (-0.25+\matrixwidth+\blockoffset+\matrixoffset*0.75,0+\matrixoffset) rectangle (-0.25+2*\matrixwidth+\blockoffset-\matrixoffset/2, \blockheight+2*\textoffset);
  \draw[draw=black, thick, rounded corners] (-0.25+2*\matrixwidth+\blockoffset+\matrixoffset/2,0+\matrixoffset) rectangle (-0.25+3*\matrixwidth+\blockoffset-\matrixoffset*0.75, \blockheight+2*\textoffset);
  \draw[draw=black, thick, rounded corners] (-0.25+3*\matrixwidth+\blockoffset+\matrixoffset*0.25,0+\matrixoffset) rectangle (-0.25+4*\matrixwidth+\blockoffset-\matrixoffset, \blockheight+2*\textoffset);
  
  \draw[draw=plotcolor6, fill=plotcolor6!20!white, thick, dashed, rounded corners=3pt] (-0.22+\blockoffset+\matrixoffset,0.03+\matrixoffset) rectangle (-0.28+\matrixwidth+\blockoffset-\matrixoffset*0.25, \blockheight+2*\textoffset-0.03);
  \draw[draw=plotcolor6, fill=plotcolor6!20!white, thick, dashed, rounded corners=3pt] (-0.22+\matrixwidth+\blockoffset+\matrixoffset*0.75,0.03+\matrixoffset) rectangle (-0.28+2*\matrixwidth+\blockoffset-\matrixoffset/2, \blockheight+2*\textoffset-0.03);
  \draw[draw=plotcolor6, fill=plotcolor6!20!white, thick, dashed, rounded corners=3pt] (-0.22+2*\matrixwidth+\blockoffset+\matrixoffset/2,0.03+\matrixoffset) rectangle (-0.28+3*\matrixwidth+\blockoffset-\matrixoffset*0.75, \blockheight+2*\textoffset-0.03);
  \draw[draw=plotcolor6, fill=plotcolor6!20!white, thick, dashed, rounded corners=3pt] (-0.22+3*\matrixwidth+\blockoffset+\matrixoffset*0.25,0.03+\matrixoffset) rectangle (-0.28+4*\matrixwidth+\blockoffset-\matrixoffset, \blockheight+2*\textoffset-0.03);

  \draw[draw=plotcolor6, thick, <-, >=stealth] (-0.25+\blockoffset+\matrixwidth*0.5, \matrixoffset) -- (0.875+\blockoffset-0.3, -0.4);
  \draw[draw=plotcolor6, thick, <-, >=stealth] (-0.25+\blockoffset+\matrixwidth*1.5, \matrixoffset) -- (0.875+\blockoffset-0.1, -0.4);
  \draw[draw=plotcolor6, thick, <-, >=stealth] (-0.25+\blockoffset+\matrixwidth*2.5, \matrixoffset) -- (0.875+\blockoffset+0.1, -0.4);
  \draw[draw=plotcolor6, thick, <-, >=stealth] (-0.25+\blockoffset+\matrixwidth*3.5, \matrixoffset) -- (0.875+\blockoffset+0.3, -0.4);
  \node[] at (0.875+\blockoffset, -0.5) {\color{plotcolor6}Per-matrix};

  \draw[draw=black, fill=white, thick] (2.625+\blockoffset, \blockheight/2) circle (0.25);
  \node[] at (2.625+\blockoffset, \blockheight/2) {\Large +};

  \draw[draw=black, fill=white, thick, rounded corners] (3.75+\blockoffset,0) rectangle (6+\blockoffset, \blockheight);
  \node[] at (4.875+\blockoffset,\blockheight+\textoffset) {MLP};
  
  \def\matrixwidth{0.75}
  \draw[draw=black, thick, thick, rounded corners] (3.75+\blockoffset+\matrixoffset,0+\matrixoffset) rectangle (3.75+\matrixwidth+\blockoffset-\matrixoffset*0.333, \blockheight+2*\textoffset);
  \draw[draw=black, thick, thick, rounded corners] (3.75+\matrixwidth+\blockoffset+\matrixoffset*0.666,0+\matrixoffset) rectangle (3.75+2*\matrixwidth+\blockoffset-\matrixoffset*0.666, \blockheight+2*\textoffset);
  \draw[draw=black, thick, thick, rounded corners] (3.75+2*\matrixwidth+\blockoffset+\matrixoffset*0.333,0+\matrixoffset) rectangle (3.75+3*\matrixwidth+\blockoffset-\matrixoffset, \blockheight+2*\textoffset);

  \draw[draw=plotcolor6, fill=plotcolor6!20!white, thick, dashed, thick, rounded corners=3pt] (3.78+\blockoffset+\matrixoffset,0.03+\matrixoffset) rectangle (3.72+\matrixwidth+\blockoffset-\matrixoffset*0.333, \blockheight+2*\textoffset-0.03);
  \draw[draw=plotcolor6, fill=plotcolor6!20!white, thick, dashed, thick, rounded corners=3pt] (3.78+\matrixwidth+\blockoffset+\matrixoffset*0.666,0.03+\matrixoffset) rectangle (3.72+2*\matrixwidth+\blockoffset-\matrixoffset*0.666, \blockheight+2*\textoffset-0.03);
  \draw[draw=plotcolor6, fill=plotcolor6!20!white, thick, dashed, thick, rounded corners=3pt] (3.78+2*\matrixwidth+\blockoffset+\matrixoffset*0.333,0.03+\matrixoffset) rectangle (3.72+3*\matrixwidth+\blockoffset-\matrixoffset, \blockheight+2*\textoffset-0.03);

  \draw[draw=plotcolor6, thick, <-, >=stealth] (3.75+\blockoffset+\matrixwidth*0.5, \matrixoffset) -- (4.875+\blockoffset-0.2, -0.4);
  \draw[draw=plotcolor6, thick, <-, >=stealth] (3.75+\blockoffset+\matrixwidth*1.5, \matrixoffset) -- (4.875+\blockoffset, -0.4);
  \draw[draw=plotcolor6, thick, <-, >=stealth] (3.75+\blockoffset+\matrixwidth*2.5, \matrixoffset) -- (4.875+\blockoffset+0.2, -0.4);
  \node[] at (4.875+\blockoffset, -0.5) {\color{plotcolor6}Per-matrix};

  \draw[draw=black, fill=white, thick] (6.625+\blockoffset, \blockheight/2) circle (0.25);
  \node[] at (6.625+\blockoffset, \blockheight/2) {\Large +};

  \draw[draw=black, thick, dotted] (-1.2+\blockoffset, \blockheight/2) -- (-0.75+\blockoffset, \blockheight/2);
  \draw[draw=black, thick, ->, >=stealth] (-0.75+\blockoffset, \blockheight/2) -- (-0.25+\blockoffset, \blockheight/2);
  \draw[draw=black, thick, ->, >=stealth] (2+\blockoffset, \blockheight/2) -- (2.375+\blockoffset, \blockheight/2);
  \draw[draw=black, thick, ->, >=stealth] (2.875+\blockoffset, \blockheight/2) -- (3.75+\blockoffset, \blockheight/2);
  \draw[draw=black, thick, ->, >=stealth] (6+\blockoffset, \blockheight/2) -- (6.375+\blockoffset, \blockheight/2);
  \draw[draw=black, thick, rounded corners, ->, >=stealth] (-0.5+\blockoffset, \blockheight/2) -- (-0.5+\blockoffset, -0.25)
                           -- (2.625+\blockoffset, -0.25) -- (2.625+\blockoffset, \blockheight/2-0.25);
  \draw[draw=black, thick, rounded corners, ->, >=stealth] (3.5+\blockoffset, \blockheight/2) -- (3.5+\blockoffset, -0.25)
                          -- (6.625+\blockoffset, -0.25) -- (6.625+\blockoffset, \blockheight/2-0.25);

  \def\blockoffset{8.5}

  \draw[draw=plotcolor4, fill=plotcolor4!20, thick, rounded corners, dashed] (-0.5+\blockoffset-2*\boxoffset,-0.75-\boxoffset) rectangle (7+\blockoffset+2*\boxoffset, \blockheight+2*\boxoffset-4*\textoffset);
  \node[] at (3.25+\blockoffset, \blockheight+1.5*\boxoffset-3*\textoffset) {\color{plotcolor4}Block size 1};
  \draw[draw=plotcolor2, fill=plotcolor2!20, thick, rounded corners, dashed] (-0.5+\blockoffset-\boxoffset,-0.75) rectangle (3+\blockoffset+\boxoffset, \blockheight+\boxoffset-2*\textoffset);
  \node[] at (1.25+\blockoffset, \blockheight+0.5*\boxoffset-\textoffset) {\color{plotcolor2}Block size $\sfrac{1}{2}$};
  \draw[draw=plotcolor2, fill=plotcolor2!20, thick, rounded corners, dashed] (3.5+\blockoffset-\boxoffset,-0.75) rectangle (7+\blockoffset+\boxoffset, \blockheight+\boxoffset-2*\textoffset);
  \node[] at (5.25+\blockoffset, \blockheight+0.5*\boxoffset-\textoffset) {\color{plotcolor2}Block size $\sfrac{1}{2}$};

  \draw[draw=black, fill=white, thick, rounded corners] (-0.25+\blockoffset,0) rectangle (2+\blockoffset, \blockheight);
  \node[] at (0.875+\blockoffset,\blockheight+\textoffset) {Attention};

  \def\matrixwidth{0.5625}
  \draw[draw=black, thick, rounded corners] (-0.25+\blockoffset+\matrixoffset,0+\matrixoffset) rectangle (-0.25+\matrixwidth+\blockoffset-\matrixoffset*0.25, \blockheight+2*\textoffset);
  \draw[draw=black, thick, rounded corners] (-0.25+\matrixwidth+\blockoffset+\matrixoffset*0.75,0+\matrixoffset) rectangle (-0.25+2*\matrixwidth+\blockoffset-\matrixoffset/2, \blockheight+2*\textoffset);
  \draw[draw=black, thick, rounded corners] (-0.25+2*\matrixwidth+\blockoffset+\matrixoffset/2,0+\matrixoffset) rectangle (-0.25+3*\matrixwidth+\blockoffset-\matrixoffset*0.75, \blockheight+2*\textoffset);
  \draw[draw=black, thick, rounded corners] (-0.25+3*\matrixwidth+\blockoffset+\matrixoffset*0.25,0+\matrixoffset) rectangle (-0.25+4*\matrixwidth+\blockoffset-\matrixoffset, \blockheight+2*\textoffset);
  
  \draw[draw=plotcolor6, fill=plotcolor6!20!white, thick, dashed, rounded corners=3pt] (-0.22+\blockoffset+\matrixoffset,0.03+\matrixoffset) rectangle (-0.28+\matrixwidth+\blockoffset-\matrixoffset*0.25, \blockheight+2*\textoffset-0.03);
  \draw[draw=plotcolor6, fill=plotcolor6!20!white, thick, dashed, rounded corners=3pt] (-0.22+\matrixwidth+\blockoffset+\matrixoffset*0.75,0.03+\matrixoffset) rectangle (-0.28+2*\matrixwidth+\blockoffset-\matrixoffset/2, \blockheight+2*\textoffset-0.03);
  \draw[draw=plotcolor6, fill=plotcolor6!20!white, thick, dashed, rounded corners=3pt] (-0.22+2*\matrixwidth+\blockoffset+\matrixoffset/2,0.03+\matrixoffset) rectangle (-0.28+3*\matrixwidth+\blockoffset-\matrixoffset*0.75, \blockheight+2*\textoffset-0.03);
  \draw[draw=plotcolor6, fill=plotcolor6!20!white, thick, dashed, rounded corners=3pt] (-0.22+3*\matrixwidth+\blockoffset+\matrixoffset*0.25,0.03+\matrixoffset) rectangle (-0.28+4*\matrixwidth+\blockoffset-\matrixoffset, \blockheight+2*\textoffset-0.03);
  \draw[draw=plotcolor6, thick, <-, >=stealth] (-0.25+\blockoffset+\matrixwidth*0.5, \matrixoffset) -- (0.875+\blockoffset-0.3, -0.4);
  \draw[draw=plotcolor6, thick, <-, >=stealth] (-0.25+\blockoffset+\matrixwidth*1.5, \matrixoffset) -- (0.875+\blockoffset-0.1, -0.4);
  \draw[draw=plotcolor6, thick, <-, >=stealth] (-0.25+\blockoffset+\matrixwidth*2.5, \matrixoffset) -- (0.875+\blockoffset+0.1, -0.4);
  \draw[draw=plotcolor6, thick, <-, >=stealth] (-0.25+\blockoffset+\matrixwidth*3.5, \matrixoffset) -- (0.875+\blockoffset+0.3, -0.4);
  \node[] at (0.875+\blockoffset, -0.5) {\color{plotcolor6}Per-matrix};
  
  \draw[draw=black, fill=white, thick] (2.625+\blockoffset, \blockheight/2) circle (0.25);
  \node[] at (2.625+\blockoffset, \blockheight/2) {\Large +};

  \draw[draw=black, fill=white, thick, rounded corners] (3.75+\blockoffset,0) rectangle (6+\blockoffset, \blockheight);
  \node[] at (4.875+\blockoffset,\blockheight+\textoffset) {MLP};
  
  \def\matrixwidth{0.75}
  \draw[draw=black, thick, thick, rounded corners] (3.75+\blockoffset+\matrixoffset,0+\matrixoffset) rectangle (3.75+\matrixwidth+\blockoffset-\matrixoffset*0.333, \blockheight+2*\textoffset);
  \draw[draw=black, thick, thick, rounded corners] (3.75+\matrixwidth+\blockoffset+\matrixoffset*0.666,0+\matrixoffset) rectangle (3.75+2*\matrixwidth+\blockoffset-\matrixoffset*0.666, \blockheight+2*\textoffset);
  \draw[draw=black, thick, thick, rounded corners] (3.75+2*\matrixwidth+\blockoffset+\matrixoffset*0.333,0+\matrixoffset) rectangle (3.75+3*\matrixwidth+\blockoffset-\matrixoffset, \blockheight+2*\textoffset);

  \draw[draw=plotcolor6, fill=plotcolor6!20!white, thick, dashed, thick, rounded corners=3pt] (3.78+\blockoffset+\matrixoffset,0.03+\matrixoffset) rectangle (3.72+\matrixwidth+\blockoffset-\matrixoffset*0.333, \blockheight+2*\textoffset-0.03);
  \draw[draw=plotcolor6, fill=plotcolor6!20!white, thick, dashed, thick, rounded corners=3pt] (3.78+\matrixwidth+\blockoffset+\matrixoffset*0.666,0.03+\matrixoffset) rectangle (3.72+2*\matrixwidth+\blockoffset-\matrixoffset*0.666, \blockheight+2*\textoffset-0.03);
  \draw[draw=plotcolor6, fill=plotcolor6!20!white, thick, dashed, thick, rounded corners=3pt] (3.78+2*\matrixwidth+\blockoffset+\matrixoffset*0.333,0.03+\matrixoffset) rectangle (3.72+3*\matrixwidth+\blockoffset-\matrixoffset, \blockheight+2*\textoffset-0.03);

  \draw[draw=plotcolor6, thick, <-, >=stealth] (3.75+\blockoffset+\matrixwidth*0.5, \matrixoffset) -- (4.875+\blockoffset-0.2, -0.4);
  \draw[draw=plotcolor6, thick, <-, >=stealth] (3.75+\blockoffset+\matrixwidth*1.5, \matrixoffset) -- (4.875+\blockoffset, -0.4);
  \draw[draw=plotcolor6, thick, <-, >=stealth] (3.75+\blockoffset+\matrixwidth*2.5, \matrixoffset) -- (4.875+\blockoffset+0.2, -0.4);
  \node[] at (4.875+\blockoffset, -0.5) {\color{plotcolor6}Per-matrix};

  \draw[draw=black, fill=white, thick] (6.625+\blockoffset, \blockheight/2) circle (0.25);
  \node[] at (6.625+\blockoffset, \blockheight/2) {\Large +};

  \draw[draw=black, thick, ->, >=stealth] (6.875, \blockheight/2) -- (-0.25+\blockoffset, \blockheight/2);
  \draw[draw=black, thick, ->, >=stealth] (2+\blockoffset, \blockheight/2) -- (2.375+\blockoffset, \blockheight/2);
  \draw[draw=black, thick, ->, >=stealth] (2.875+\blockoffset, \blockheight/2) -- (3.75+\blockoffset, \blockheight/2);
  \draw[draw=black, thick, ->, >=stealth] (6+\blockoffset, \blockheight/2) -- (6.375+\blockoffset, \blockheight/2);
  \draw[draw=black, thick, rounded corners, ->, >=stealth] (-0.5+\blockoffset, \blockheight/2) -- (-0.5+\blockoffset, -0.25)
                           -- (2.625+\blockoffset, -0.25) -- (2.625+\blockoffset, \blockheight/2-0.25);
  \draw[draw=black, thick, rounded corners, ->, >=stealth] (3.5+\blockoffset, \blockheight/2) -- (3.5+\blockoffset, -0.25)
                          -- (6.625+\blockoffset, -0.25) -- (6.625+\blockoffset, \blockheight/2-0.25);
  \draw[draw=black, thick] (6.875+\blockoffset, \blockheight/2) -- (7.25+\blockoffset, \blockheight/2);
  \draw[draw=black, thick, dotted] (7.25+\blockoffset, \blockheight/2) -- (7.7+\blockoffset, \blockheight/2);
\end{tikzpicture}

%% file: ppl_vs_mem_bar_by_model.tex
\centering
\small

\edef\barwidth{5.5pt}
\edef\barshift{8pt}

\begin{tikzpicture}
\hspace{-5pt}
    \begin{groupplot}[
        group style={
            group size=3 by 1,
            horizontal sep=0.82cm,
            vertical sep=2cm,
        },
        width=0.392\textwidth,
        height=0.32\textwidth,
        legend columns=4,
        grid=major,
        cycle list name=tab10colors,
        table/col sep=comma,
        every axis plot/.append style={fill opacity=0.4},
        enlarge x limits=0.25,
    ]

    \nextgroupplot[title={Perplexity}, ylabel={},
                   xlabel={},
                   xlabel style={
                       at={(axis description cs:0.5,-0.12)},
                       anchor=north,
                       rotate=0
                   },
                   xtick=data,
                   xticklabels={{\footnotesize LLaMA-3\\\footnotesize8B},{\footnotesize LLaMA-2\\\footnotesize13B},{\footnotesize Qwen-2.5\\\footnotesize32B}},
                   ymin=0,
                   xtick style={draw=none},
                   xticklabel style={align=center},
                   yticklabel style={font=\footnotesize},
                   ]
        \addplot+[fill, ybar, bar width=\barwidth, bar shift=-1.5*\barshift, mark=none]
                    table[
                        x expr=\coordindex,
                        y=-1,
                        col sep=comma
                    ]{data/ppl_vs_mem/ppl_by_model.txt};
        \addplot+[fill, ybar, bar width=\barwidth,  bar shift=-0.5*\barshift, mark=none]
                    table[
                        x expr=\coordindex,
                        y=1,
                        col sep=comma
                    ]{data/ppl_vs_mem/ppl_by_model.txt};
        \addplot+[fill, ybar, bar width=\barwidth, bar shift=0.5*\barshift, mark=none]
                    table[
                        x expr=\coordindex,
                        y=2,
                        col sep=comma
                    ]{data/ppl_vs_mem/ppl_by_model.txt};
        \addplot+[fill, ybar, bar width=\barwidth, bar shift=1.5*\barshift, mark=none]
                    table[
                        x expr=\coordindex,
                        y=6,
                        col sep=comma
                    ]{data/ppl_vs_mem/ppl_by_model.txt};

    \nextgroupplot[title={Peak GPU memory usage (GB)}, ylabel={},
                   xlabel={},
                   xlabel style={
                       at={(axis description cs:0.5,-0.12)},
                       anchor=north,
                       rotate=0
                   },
                   xtick=data,
                   xticklabels={{\footnotesize LLaMA-3\\\footnotesize8B},{\footnotesize LLaMA-2\\\footnotesize13B},{\footnotesize Qwen-2.5\\\footnotesize32B}},
                   ymin=0,
                   xtick style={draw=none},
                   xticklabel style={align=center},
                   yticklabel style={font=\footnotesize},
                   ]
        \addplot+[fill, ybar, bar width=\barwidth, bar shift=-1.5*\barshift, mark=none]
                    table[
                        x expr=\coordindex,
                        y=-1,
                        col sep=comma
                    ]{data/ppl_vs_mem/mem_by_model.txt};
        \addplot+[fill, ybar, bar width=\barwidth, bar shift=-0.5*\barshift, mark=none]
                    table[
                        x expr=\coordindex,
                        y=1,
                        col sep=comma
                    ]{data/ppl_vs_mem/mem_by_model.txt};
        \addplot+[fill, ybar, bar width=\barwidth, bar shift=0.5*\barshift, mark=none]
                    table[
                        x expr=\coordindex,
                        y=2,
                        col sep=comma
                    ]{data/ppl_vs_mem/mem_by_model.txt};
        \addplot+[fill, ybar, bar width=\barwidth, bar shift=1.5*\barshift, mark=none]
                    table[
                        x expr=\coordindex,
                        y=6,
                        col sep=comma
                    ]{data/ppl_vs_mem/mem_by_model.txt};

        \nextgroupplot[title={Compute time (minutes)}, ylabel={},
                   xlabel={},
                   xlabel style={
                       at={(axis description cs:0.5,-0.12)},
                       anchor=north,
                       rotate=0
                   },
                   xtick=data,
                   xticklabels={{\footnotesize LLaMA-3\\\footnotesize8B},{\footnotesize LLaMA-2\\\footnotesize13B},{\footnotesize Qwen-2.5\\\footnotesize32B}},
                   ymin=0,
                   xtick style={draw=none},
                   xticklabel style={align=center},
                   yticklabel style={font=\footnotesize},
                   legend style={{at=(-1.95, -0.35)}, anchor=north west, /tikz/column sep=3pt},
        legend image code/.code={\draw[#1, fill, draw, fill opacity=0.4] (0cm,-0.1cm) rectangle (0.2cm,0.1cm);},
                   ]
        \addplot+[fill, ybar, bar width=\barwidth, bar shift=-1.5*\barshift, mark=none]
                    table[
                        x expr=\coordindex,
                        y=-1,
                        col sep=comma
                    ]{data/ppl_vs_mem/time_by_model.txt};
        \addlegendentry{\footnotesize Attention and MLP separately}
        \addplot+[fill, ybar, bar width=\barwidth, bar shift=-0.5*\barshift, mark=none]
                    table[
                        x expr=\coordindex,
                        y=1,
                        col sep=comma
                    ]{data/ppl_vs_mem/time_by_model.txt};
        \addlegendentry{\footnotesize Full block}
        \addplot+[fill, ybar, bar width=\barwidth, bar shift=0.5*\barshift, mark=none]
                    table[
                        x expr=\coordindex,
                        y=2,
                        col sep=comma
                    ]{data/ppl_vs_mem/time_by_model.txt};
        \addlegendentry{\footnotesize Two blocks}
        \addplot+[fill, ybar, bar width=\barwidth, bar shift=1.5*\barshift, mark=none]
                    table[
                        x expr=\coordindex,
                        y=6,
                        col sep=comma
                    ]{data/ppl_vs_mem/time_by_model.txt};
        \addlegendentry{\footnotesize $\sfrac{1}{4}$ model}
    \end{groupplot}
\end{tikzpicture}
\normalsize

%% file: ppl_vs_numsamples_only.tex
\centering
\scriptsize

\begin{tikzpicture}
    \begin{groupplot}[
        group style={
            group size=1 by 1,
            horizontal sep=0.5cm,
            vertical sep=1.5cm,
        },
        width=1.114\textwidth,
        height=0.7\textwidth,
        error bars/y dir=both,
        error bars/y explicit,
        grid=major,
        cycle list name=tab10colors,
        ymin=15.1,ymax=17.7,
        enlarge x limits=0.04,
    ]

    \nextgroupplot[align=center, ylabel={\scriptsize Perplexity},
                   ylabel style={
                       at={(axis description cs:-0.06,0.5)},
                   }, xlabel={\scriptsize Number of samples},
                   xmode=log,
                   xtick={10,100,1000,10000, 100000, 1000000},
                   xticklabels={10,100,1k,10k, 100k, 1M},
                   legend style={{at=(0.0, 1.05)}, anchor=south west},
                   legend columns=3,
                   ]
    \addplot[name path=max_0_5, draw=none, forget plot]
      table[x=reconstruct_n_samples, y=ppl_-1, col sep=comma]
      {data/ppl_by_numsamples/OPT_1_3B_Wanda/Calibration_set_max.csv};
    \addplot[name path=min_0_5, draw=none, forget plot]
      table[x=reconstruct_n_samples, y=ppl_-1, col sep=comma]
    {data/ppl_by_numsamples/OPT_1_3B_Wanda/Calibration_set_min.csv};
    \addplot[fill=plotcolor1, fill opacity=0.4, forget plot]
      fill between[of=max_0_5 and min_0_5];

    \addplot[name path=max_1, draw=none, forget plot]
      table[x=reconstruct_n_samples, y=ppl_1, col sep=comma]
      {data/ppl_by_numsamples/OPT_1_3B_Wanda/Calibration_set_max.csv};
    \addplot[name path=min_1, draw=none, forget plot]
      table[x=reconstruct_n_samples, y=ppl_1, col sep=comma]
    {data/ppl_by_numsamples/OPT_1_3B_Wanda/Calibration_set_min.csv};
    \addplot[fill=plotcolor2, fill opacity=0.4, forget plot]
      fill between[of=max_1 and min_1];

    \addplot[name path=max_12, draw=none, forget plot]
      table[x=reconstruct_n_samples, y=ppl_12, col sep=comma]
      {data/ppl_by_numsamples/OPT_1_3B_Wanda/Calibration_set_max.csv};
    \addplot[name path=min_12, draw=none, forget plot]
      table[x=reconstruct_n_samples, y=ppl_12, col sep=comma]
    {data/ppl_by_numsamples/OPT_1_3B_Wanda/Calibration_set_min.csv};
    \addplot[fill=plotcolor3, fill opacity=0.4, forget plot]
      fill between[of=max_12 and min_12];

    \addplot[name path=max_24, draw=none, forget plot]
      table[x=reconstruct_n_samples, y=ppl_24, col sep=comma]
      {data/ppl_by_numsamples/OPT_1_3B_Wanda/Calibration_set_max.csv};
    \addplot[name path=min_24, draw=none, forget plot]
      table[x=reconstruct_n_samples, y=ppl_24, col sep=comma]
    {data/ppl_by_numsamples/OPT_1_3B_Wanda/Calibration_set_min.csv};
    \addplot[fill=plotcolor6, fill opacity=0.4, forget plot]
      fill between[of=max_24 and min_24];

    \addplot[name path=max_ft, draw=none, forget plot]
      table[x=reconstruct_n_samples, y=ppl_1, col sep=comma]
      {data/ppl_by_numsamples/OPT_1_3B_Wanda/Ret_Calibration_set_max.csv};
    \addplot[name path=min_ft, draw=none, forget plot]
      table[x=reconstruct_n_samples, y=ppl_1, col sep=comma]
    {data/ppl_by_numsamples/OPT_1_3B_Wanda/Ret_Calibration_set_min.csv};
    \addplot[fill=plotcolor5, fill opacity=0.4, forget plot]
      fill between[of=max_ft and min_ft];

    \addplot+[mark=none, mark size=1pt, color=plotcolor1, line width=0.6pt] table[
                x=reconstruct_n_samples,
                y=ppl_-1,
                col sep=comma
        ]{data/ppl_by_numsamples/OPT_1_3B_Wanda/Calibration_set_mean.csv};
    \addlegendentry{\scriptsize Block size $\sfrac{1}{2}$}
    \addplot+[mark=none, mark size=1pt, color=plotcolor2, line width=0.6pt] table[
                x=reconstruct_n_samples,
                y=ppl_1,
                col sep=comma
        ]{data/ppl_by_numsamples/OPT_1_3B_Wanda/Calibration_set_mean.csv};
    \addlegendentry{\scriptsize Block size 1}
    \addplot+[mark=none, mark size=1pt, color=plotcolor3, line width=0.6pt] table[
                x=reconstruct_n_samples,
                y=ppl_12,
                col sep=comma
        ]{data/ppl_by_numsamples/OPT_1_3B_Wanda/Calibration_set_mean.csv};
    \addlegendentry{\scriptsize Block size 12}
    \addplot+[mark=none, mark size=1pt, color=plotcolor6, line width=0.6pt] table[
                x=reconstruct_n_samples,
                y=ppl_24,
                col sep=comma
        ]{data/ppl_by_numsamples/OPT_1_3B_Wanda/Calibration_set_mean.csv};
    \addlegendentry{\scriptsize Full decoder}
    \addplot+[mark=none, mark size=1pt, color=plotcolor5, line width=0.6pt] table[
                x=reconstruct_n_samples,
                y=ppl_1,
                col sep=comma
        ]{data/ppl_by_numsamples/OPT_1_3B_Wanda/Ret_Calibration_set_mean.csv};
    \addlegendentry{\scriptsize Full fine-tuning}

    \end{groupplot}
\end{tikzpicture}
\normalsize

%% file: lr_figure2.tex
\centering
\scriptsize

\begin{tikzpicture}
    \begin{groupplot}[
        group style={
            group size=1 by 1,
            horizontal sep=0.5cm,
            vertical sep=1cm,
        },
        width=1.09\textwidth,
        height=0.7\textwidth,
        grid=major,
        cycle list name=tab10colors,
        ymin=30,ymax=66,
        enlarge x limits=0.03,
    ]

    \nextgroupplot[align=center, ylabel={Accuracy (in \%)},
                   ylabel style={
                       at={(axis description cs:-0.07,0.5)},
                   }, xlabel={Learning rate},
                   legend style={{at=(0.0, 1.05)}, anchor=south west, column sep=1.4pt},
                   legend columns=3,
                   xmode=log,
                   ]

    \addplot[name path=max_bs_1_2, draw=none, forget plot]
      table[x=lr, y=-1, col sep=comma]
      {data/lr_ablation/qwen7granularities_max.txt};
    \addplot[name path=min_bs_1_2, draw=none, forget plot]
      table[x=lr, y=-1, col sep=comma]
      {data/lr_ablation/qwen7granularities_min.txt};
    \addplot[fill=plotcolor1, fill opacity=0.4, forget plot]
      fill between[of=max_bs_1_2 and min_bs_1_2];

      \addplot[name path=max_bs_1, draw=none, forget plot]
      table[x=lr, y=1, col sep=comma]
      {data/lr_ablation/qwen7granularities_max.txt};
    \addplot[name path=min_bs_1, draw=none, forget plot]
      table[x=lr, y=1, col sep=comma]
      {data/lr_ablation/qwen7granularities_min.txt};
    \addplot[fill=plotcolor2, fill opacity=0.4, forget plot]
      fill between[of=max_bs_1 and min_bs_1];

      \addplot[name path=max_bs_2, draw=none, forget plot]
      table[x=lr, y=2, col sep=comma]
      {data/lr_ablation/qwen7granularities_max.txt};
    \addplot[name path=min_bs_2, draw=none, forget plot]
      table[x=lr, y=2, col sep=comma]
      {data/lr_ablation/qwen7granularities_min.txt};
    \addplot[fill=plotcolor3, fill opacity=0.4, forget plot]
      fill between[of=max_bs_2 and min_bs_2];

      \addplot[name path=max_bs_7, draw=none, forget plot]
      table[x=lr, y=7, col sep=comma]
      {data/lr_ablation/qwen7granularities_max.txt};
    \addplot[name path=min_bs_7, draw=none, forget plot]
      table[x=lr, y=7, col sep=comma]
      {data/lr_ablation/qwen7granularities_min.txt};
    \addplot[fill=plotcolor5, fill opacity=0.4, forget plot]
      fill between[of=max_bs_7 and min_bs_7];

    \addplot+[mark=none, mark size=1pt, color=plotcolor1, line width=0.6pt] table[
                x=lr,
                y=-1,
                col sep=comma
        ]{data/lr_ablation/qwen7granularities.txt};
    \addlegendentry{\scriptsize Block size $\sfrac{1}{2}$}
    \addplot+[mark=none, mark size=1pt, color=plotcolor2, line width=0.6pt] table[
                x=lr,
                y=1,
                col sep=comma
        ]{data/lr_ablation/qwen7granularities.txt};
    \addlegendentry{\scriptsize Block size $1$}
    \addplot+[mark=none, mark size=1pt, color=plotcolor3, line width=0.6pt] table[
                x=lr,
                y=2,
                col sep=comma
        ]{data/lr_ablation/qwen7granularities.txt};
    \addlegendentry{\scriptsize Block size $2$}
    \addplot+[mark=none, mark size=1pt, color=plotcolor5, line width=0.6pt] table[
                x=lr,
                y=7,
                col sep=comma
        ]{data/lr_ablation/qwen7granularities.txt};
    \addlegendentry{\scriptsize Block size $7$}
    \end{groupplot}
\end{tikzpicture}
\normalsize
\vspace{-1pt}

%% file: error_plot.tex
\centering
\scriptsize

\begin{tikzpicture}
    \begin{groupplot}[
        group style={
            group size=1 by 1,
            horizontal sep=0.5cm,
            vertical sep=1cm,
        },
        width=1.09\textwidth,
        height=0.7\textwidth,
        grid=major,
        cycle list name=tab10colors,
        enlarge x limits=0.03,
    ]

    \nextgroupplot[align=center, ylabel={Normalized MSE},
                   ylabel style={
                       at={(axis description cs:-0.095,0.5)},
                   }, xlabel={Transformer block index},
                   xlabel style={
                       at={(axis description cs:0.5,-0.12)},
                       anchor=north,
                       rotate=0
                   },
                   legend style={{at=(0.0, 1.05)}, anchor=south west, column sep=1.4pt},
                   legend columns=3,
                   yticklabel style={font=\scriptsize},
                   ]
    
    \addplot+[mark=none, mark size=1pt, color=plotcolor1, line width=0.6pt] table[
                x=layer,
                y=-2,
                col sep=comma
            ]{data/error_plot/errors_k_proj.csv};
    \addlegendentry{\scriptsize Per-matrix}
    \addplot+[mark=none, mark size=1pt, color=plotcolor2, line width=0.6pt] table[
                x=layer,
                y=-1,
                col sep=comma
        ]{data/error_plot/errors_k_proj.csv};
    \addlegendentry{\scriptsize Block size $\sfrac{1}{2}$}
    \addplot+[mark=none, mark size=1pt, color=plotcolor3, line width=0.6pt] table[
                x=layer,
                y=1,
                col sep=comma
        ]{data/error_plot/errors_k_proj.csv};
    \addlegendentry{\scriptsize Block size $1$}
    \addplot+[mark=none, mark size=1pt, color=plotcolor6, line width=0.6pt] table[
                x=layer,
                y=2,
                col sep=comma
        ]{data/error_plot/errors_k_proj.csv};
    \addlegendentry{\scriptsize Block size $2$}
    \addplot+[mark=none, mark size=1pt, color=plotcolor5, line width=0.6pt] table[
                x=layer,
                y=4,
                col sep=comma
        ]{data/error_plot/errors_k_proj.csv};
    \addlegendentry{\scriptsize Block size $4$}
    \end{groupplot}
\end{tikzpicture}
\normalsize

%% file: error_plot_local.tex
\centering
\scriptsize

\begin{tikzpicture}
    \begin{groupplot}[
        group style={
            group size=1 by 1,
            horizontal sep=0.5cm,
            vertical sep=1cm,
        },
        width=1.09\textwidth,
        height=0.7\textwidth,
        grid=major,
        cycle list name=tab10colors,
        enlarge x limits=0.03,
    ]

    \nextgroupplot[align=center, ylabel={Normalized MSE},
                   ylabel style={
                       at={(axis description cs:-0.095,0.5)},
                   }, xlabel={Transformer block index},
                   xlabel style={
                       at={(axis description cs:0.5,-0.12)},
                       anchor=north,
                       rotate=0
                   },
                   legend style={{at=(0.0, 1.05)}, anchor=south west, column sep=1.4pt},
                   legend columns=3,
                   yticklabel style={font=\scriptsize},
                   every y tick scale label/.style={
                        at={(-0.15, 1.0)},
                        anchor=north west,
                    },
                   ]
    
    \addplot+[mark=none, mark size=1pt, color=plotcolor1, line width=0.6pt] table[
                x=layer,
                y=-2,
                col sep=comma
            ]{data/error_plot/local_errors_k_proj.csv};
    \addlegendentry{\scriptsize Per-matrix}
    \addplot+[mark=none, mark size=1pt, color=plotcolor2, line width=0.6pt] table[
                x=layer,
                y=-1,
                col sep=comma
        ]{data/error_plot/local_errors_k_proj.csv};
    \addlegendentry{\scriptsize Block size $\sfrac{1}{2}$}
    \addplot+[mark=none, mark size=1pt, color=plotcolor3, line width=0.6pt] table[
                x=layer,
                y=1,
                col sep=comma
        ]{data/error_plot/local_errors_k_proj.csv};
    \addlegendentry{\scriptsize Block size $1$}
    \addplot+[mark=none, mark size=1pt, color=plotcolor6, line width=0.6pt] table[
                x=layer,
                y=2,
                col sep=comma
        ]{data/error_plot/local_errors_k_proj.csv};
    \addlegendentry{\scriptsize Block size $2$}
    \addplot+[mark=none, mark size=1pt, color=plotcolor5, line width=0.6pt] table[
                x=layer,
                y=4,
                col sep=comma
        ]{data/error_plot/local_errors_k_proj.csv};
    \addlegendentry{\scriptsize Block size $4$}
    \end{groupplot}
\end{tikzpicture}
\normalsize

%% file: error_hist.tex
\centering
\small

\edef\barwidth{3pt}
\edef\barshift{8pt}

\begin{tikzpicture}
\hspace{-5pt}
    \begin{groupplot}[
        group style={
            group size=1 by 1,
            horizontal sep=0.5cm,
            vertical sep=2cm,
        },
        width=1.091\textwidth,
        height=0.7\textwidth,
        legend columns=2,
        grid=major,
        cycle list name=tab10colors,
        table/col sep=comma,
        xtick style={draw=none},
        xticklabel style={align=center, font=\scriptsize},
        every axis plot/.append style={fill opacity=0.4},
        enlarge x limits=0.05,
    ]

        \nextgroupplot[ylabel={\scriptsize Normalized count},
                   xlabel={\scriptsize Normalized distance},
                   xlabel style={
                       at={(axis description cs:0.5,-0.12)},
                       anchor=north,
                       rotate=0
                   },
                   ymin=0,
                   xtick={0, 10, 20, 30},
                   xticklabels={0.00, 0.26, 0.52, 0.78},
                   yticklabel style={font=\scriptsize},
                   legend style={{at=(0.0, 1.05)}, anchor=south west, column sep=3.11pt, inner ysep=7pt, inner xsep=11.7pt},
        legend image code/.code={\draw[#1, fill, draw, fill opacity=0.4] (0cm,-0.1cm) rectangle (0.2cm,0.1cm);},
                   ]
        \addplot+[fill, ybar, bar width=\barwidth, mark=none, color=plotcolor2]
                    table[
                        x expr=\coordindex,
                        y=hist_other,
                        col sep=comma
                    ]{data/error_hist/error_hist_k_proj.csv};
        \addlegendentry{\scriptsize Coarse to coarse\ \ \ \ \ }
        \addplot+[fill, ybar, bar width=\barwidth, mark=none, color=plotcolor5]
                    table[
                        x expr=\coordindex,
                        y=hist_pm,
                        col sep=comma
                    ]{data/error_hist/error_hist_k_proj.csv};
        \addlegendentry{\scriptsize Per-matrix to coarse}
    \end{groupplot}
\end{tikzpicture}
\normalsize
\vspace{-1pt}

%% file: acc_vs_scale.tex
\centering
\scriptsize

\begin{tikzpicture}
    \begin{groupplot}[
        group style={
            group size=1 by 1,
            horizontal sep=0.5cm,
            vertical sep=1cm,
        },
        width=1.1115\textwidth,
        height=0.7\textwidth,
        grid=major,
        cycle list name=tab10colors,
        enlarge x limits=0.03,
    ]

    \nextgroupplot[align=center, ylabel={Error rate (in \%)},
                   ylabel style={
                       at={(axis description cs:-0.06,0.5)},
                   }, xlabel={Model size},
                   legend style={{at=(0.0, 1.05)}, anchor=south west, column sep=0.0pt, inner xsep=1pt},
                   legend columns=3,
                   xmode=log,
                   ymode=log,
                   xtick={500000000, 3000000000, 7000000000, 14000000000, 32000000000, 72000000000},
                   xticklabels={500M, 3B, 7B, 14B, 32B, 72B},
                   ytick={30, 40, 50, 60, 70},
                   yticklabels={30, 40, 50, 60, 70},
                   ]
      
    \addplot+[mark=none, mark size=1pt, color=plotcolor1, line width=1pt, dashed] table[
                x=param,
                y expr={100 - \thisrow{magnorec}},
                col sep=comma
        ]{data/acc_vs_scale_2_4.csv};
    \addlegendentry{ Magnitude}

    \addplot+[mark=none, mark size=1pt, color=plotcolor2, line width=1pt, dashed] table[
                x=param,
                y expr={100 - \thisrow{wandanorec}},
                col sep=comma
        ]{data/acc_vs_scale_2_4.csv};
    \addlegendentry{ Wanda}

    \addplot+[mark=none, mark size=1pt, color=plotcolor5, line width=1pt, dashed] table[
                x=param,
                y expr={100 - \thisrow{sgptnorec}},
                col sep=comma
        ]{data/acc_vs_scale_2_4.csv};
    \addlegendentry{ SparseGPT}

    \addplot+[mark=none, mark size=1pt, color=plotcolor1, line width=1pt] table[
                x=param,
                y expr={100 - \thisrow{magrec}},
                col sep=comma
        ]{data/acc_vs_scale_2_4.csv};
    \addlegendentry{ Magnitude+rec.}
    
    \addplot+[mark=none, mark size=1pt, color=plotcolor2, line width=1pt] table[
                x=param,
                y expr={100 - \thisrow{wandarec}},
                col sep=comma
        ]{data/acc_vs_scale_2_4.csv};
    \addlegendentry{ Wanda+rec.}
    
    \addplot+[mark=none, mark size=1pt, color=plotcolor5, line width=1pt] table[
                x=param,
                y expr={100 - \thisrow{sgptrec}},
                col sep=comma
        ]{data/acc_vs_scale_2_4.csv};
    \addlegendentry{ SparseGPT+rec.}
    \end{groupplot}
\end{tikzpicture}
\vspace{-9.1pt}
\normalsize